\documentclass[sigconf, nonacm]{acmart}
\usepackage{balance}
\usepackage[normalem]{ulem}
\usepackage{color,soul}
\newcommand\vldbdoi{XX.XX/XXX.XX}
\newcommand\vldbpages{XXX-XXX}
\newcommand\vldbvolume{14}
\newcommand\vldbissue{1}
\newcommand\vldbyear{2020}
\newcommand\vldbauthors{\authors}
\newcommand\vldbtitle{\shorttitle} 
\newcommand\vldbavailabilityurl{http://vldb.org/pvldb/format_vol14.html}
\newcommand\vldbpagestyle{plain} 

\usepackage{listings}
\usepackage{balance}
\usepackage[a-1b]{pdfx}
\definecolor{darkgreen}{rgb}{0.0, 0.2, 0.13}
\definecolor{celadon}{rgb}{0.67, 0.88, 0.69}
\definecolor{forestgreen(traditional)}{rgb}{0.0, 0.27, 0.13}
\definecolor{forestgreen(web)}{rgb}{0.13, 0.55, 0.13}

\begin{document}
\title{Large Graph Convolutional Network Training with GPU-Oriented Data Communication Architecture}

%%
%% The "author" command and its associated commands are used to define the authors and their affiliations.
\author{Seung Won Min}
\affiliation{%
  \institution{UIUC}
}
\email{min16@illinois.edu}

\author{Kun Wu}
\affiliation{%
  \institution{UIUC}
}
\email{kunwu2@illinois.edu}

\author{Sitao Huang}
\affiliation{%
  \institution{UIUC}
}
\email{shuang91@illinois.edu}

\author{Mert Hidayetoğlu}
\affiliation{%
  \institution{UIUC}
}
\email{hidayet2@illinois.edu}

\author{Jinjun Xiong}
\affiliation{%
  \institution{IBM T.J. Watson Research Center}
}
\email{jinjun@us.ibm.com}

\author{Eiman Ebrahimi}
\affiliation{%
  \institution{NVIDIA}
}
\email{eebrahimi@nvidia.com}

\author{Deming Chen}
\affiliation{%
  \institution{UIUC}
}
\email{dchen@illinois.edu}

\author{Wen-mei Hwu}
\affiliation{%
  \institution{NVIDIA / UIUC}
}
\email{whwu@nvidia.com}

\ifdefined\submit
\newcommand{\david}[1]{{}}
\newcommand{\kun}[1]{}
\newcommand{\sh}[1]{{}}
\newcommand{\mert}[1]{{}}
\newcommand{\jx}[1]{{}}
\newcommand{\dc}[1]{{}}
\newcommand{\wmh}[1]{}
\newcommand{\eiman}[1]{{}}
\newcommand{\kunhl}[1]{{#1}}
\else
\newcommand{\david}[1]{[{\color{orange}David: #1}]}
\newcommand{\kun}[1]{[{\color{blue}Kun: #1}]}
\newcommand{\kunhl}[1]{{\hl{#1}}}
\newcommand{\sh}[1]{[{\color{purple}SH: #1}]}
\newcommand{\mert}[1]{[{\color{red}Mert: #1}]}
\newcommand{\jx}[1]{[{\color{red}JX: #1}]}
\newcommand{\dc}[1]{[{\color{red}DC: #1}]}
\newcommand{\wmh}[1]{[{\color{red}WMH: #1}]}
\newcommand{\eiman}[1]{[{\color{red}Eiman: #1}]}
\fi

\begin{abstract}

%The growing adoption of 
Graph Convolutional Networks (GCNs) are increasingly adopted in large-scale graph-based recommender systems.
%Graph Convolutional Networks (GCNs) \jx{are} \jx{remove: have been} increasingly adopted in large-scale graph-based recommender system. %has motivated deep learning frameworks to support GCN training. 
%A GCN consists of an embedding vector generator and a neural network.
Training GCN requires the minibatch generator traversing graphs and sampling the sparsely located neighboring nodes to obtain their features.
%\jx{Training}
%\jx{remove: Unlike the purely content based neural networks, training} GCN requires the minibatch generator \jx{traversing} \jx{remove: to traverse} graphs and \jx{sampling} \jx{remove: to sample} the
%\jx{sparse} neighboring nodes \jx{to obtain
%their features of interests}.
%
%\jx{remove: The sparse node features of interest are then used to train the neural network.}
%
%
Since real-world graphs often exceed the capacity of GPU memory, current GCN training systems keep the feature table in host memory and rely on the CPU to collect sparse features before sending them to the GPUs.
This approach, however, puts tremendous pressure on host memory bandwidth and the CPU. 
%\jx{Since} real-world graphs often exceed the capacity of GPU memory, {\jx as a result}, current \jx{GCN training} systems keep their inputs in the host memory, \jx{and} use the CPU to collect the sparse features \jx{remove:,} and transfer the features to the GPUs for GCN training. 
%\jx{This approach, however, puts tremendous pressure on the host memory bandwidth and CPU}
%\jx{remove:
%However, this approach makes excessive use of the host memory bandwidth} 
This is because the CPU needs to (1) read sparse features from memory, (2) write features into memory as a dense format, and (3) transfer the features from memory to the GPUs.
%
%\david{This could be misleading because what we really doing is reorganizing (gathering) the already existing features for better DMA. CPU is not really generating the features. Figure~\ref{fig:combined} (a)} 

In this work, we propose a novel GPU-oriented data communication approach for GCN training, where 
%the GPU generates the feature vectors by directly reading the node attributes from the host memory. This direct approach eliminates the majority of the host memory accesses in current systems. %Unlike the conventional block data transfer approaches, Our method uses 
%the CPU only sends the indices of the sparse features to the GPUs and let
GPU threads directly access sparse features in host memory through zero-copy accesses without much CPU help.
%the CPU traverses the graph, generates the indices of all the sampled neighbor nodes, and transfers the indices to the GPU. These indices are used by GPU threads to access the attribute values in the host memory through the zero-copy access hardware support, and collect them into embedding vectors in the GPU memory. 
%which can access complicated data structures like graphs over the external interconnects efficiently.
%This approach eliminates the extra steps that write the features vectors into the host memory and read these feature vectors for transfer to the GPU memory. 
%Since the size of embedding vectors are much larger than that of index values, our method significantly reduces the consumption of the host memory bandwidth.
By removing the CPU gathering stage, our method significantly reduces the consumption of the host resources and data access latency.
We further present two important techniques to achieve high host memory access efficiency by the GPU: (1) automatic data access address alignment to maximize PCIe packet efficiency, and (2) asynchronous zero-copy access and kernel execution to fully overlap data transfer with training.
%GPU-oriented data communication architecture to best utilize the slow external interconnect during GCN training.
%
%
%Without the need of creating blocks for the data transfers, our method not only increases the data transfer efficiency but also greatly reduces the CPU and memory utilization.
%
%Our work allows also introduces automatic data access alignment and asynchronous zero-copy operation optimizations to provide the maximum data transfer efficiency throughout the training time.
%
% deploy our work through PyTorch, which is one of the most popular python-based deep learning libraries. 
%Our system-level modifications are incorporated into the PyTorch source. code to shield the general users from the low-level details.
%To minimize the changes in existing PyTorch programming model, most of our implementation reuses the current PyTorch functionality.
%
%Other system-level modifications are embedded into PyTorch source code to avoid general users from dealing with low-level programmings.
%
We incorporate our method into PyTorch and evaluate its effectiveness using several graphs with sizes up to 111 million nodes and 1.6 billion edges.
%
%\jx{Comments: Do we have single-GPU result as well? It's good to mention the single-GPU results first before we move to multi-GPUs.}
In a multi-GPU training setup, our method is 65-92\% faster than the conventional data transfer method, and can even match the performance of all-in-GPU-memory training for some graphs that fit in GPU memory. 
%\dc{I think abstract can be shortened to make sure the key ideas/contributions would stand out clearly.} \kun{Seems to me smoother while retaining all the major points by removing the second to the last sentence in the first paragraph.}

\end{abstract}

\maketitle

\pagestyle{\vldbpagestyle}
\begingroup\small\noindent\raggedright\textbf{PVLDB Reference Format:}\\
\vldbauthors. \vldbtitle. PVLDB, \vldbvolume(\vldbissue): \vldbpages, \vldbyear.\\
\href{https://doi.org/\vldbdoi}{doi:\vldbdoi}
\endgroup
\begingroup
\renewcommand\thefootnote{}\footnote{\noindent
This work is licensed under the Creative Commons BY-NC-ND 4.0 International License. Visit \url{https://creativecommons.org/licenses/by-nc-nd/4.0/} to view a copy of this license. For any use beyond those covered by this license, obtain permission by emailing \href{mailto:info@vldb.org}{info@vldb.org}. Copyright is held by the owner/author(s). Publication rights licensed to the VLDB Endowment. \\
\raggedright Proceedings of the VLDB Endowment, Vol. \vldbvolume, No. \vldbissue\ %
ISSN 2150-8097. \\
\href{https://doi.org/\vldbdoi}{doi:\vldbdoi} \\
}\addtocounter{footnote}{-1}\endgroup

\ifdefempty{\vldbavailabilityurl}{}{
\vspace{.3cm}
\begingroup\small\noindent\raggedright\textbf{PVLDB Artifact Availability:}\\
The source code, data, and/or other artifacts have been made available at \url{https://github.com/K-Wu/pytorch-direct_dgl}.
\endgroup
}

\section{Introduction}

Acceleration of modern machine learning models is often severely limited by insufficient memory bandwidth~\cite{8114708,strom2015scalable,lin2018deep}.
To provide the best possible memory bandwidth, data is usually placed in
%\kun{\sout{is located to} are placed in}
memory closest to the processing units of the accelerators~\cite{10.1145/3200691.3178491,7783721}.
However, with extremely large datasets, it is inevitable to put
%\kun{\sout{locate} put}
data farther from the processing units to take advantage of larger capacity (e.g., host memory).
In this case directly accessing remote data from the processing units can be very inefficient due to slow external interconnects.
%In this case, due to the slow external interconnects, it can be very inefficient to use processing units to access remote data by themselves.
%
To free processing units from spending excessive amount of time accessing remote data, modern hardware systems utilize direct memory access (DMA) engines.

\begin{figure}[t]
  \centering
  \includegraphics[width=\linewidth]{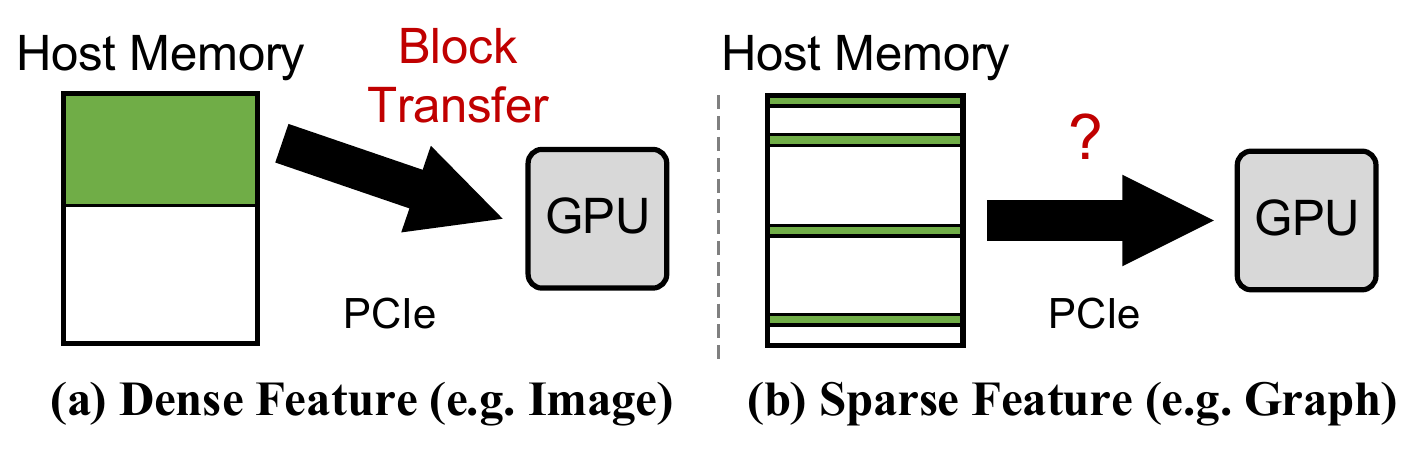}
  \caption{Challenge of GPUs accessing fine-grained sparse features in host memory.}
  \label{fig:dense_vs_sparse}
\end{figure}

DMA engines are specialized in transferring large blocks of data independently.
By providing source and destination memory pointers along with the data size, DMA engines transfer data behind the scenes while keeping
%\kun{\sout{leaving} setting free}
processing units available for other tasks.
Initiating each DMA requires multiple interactions between the user application and the operating system, but these overheads can be offset by transferring large data blocks (Figure~\ref{fig:dense_vs_sparse} (a)).%increasing the data transfer size per DMA.

The recent adaptation of machine learning to a wide range of tasks has led modern deep neural networks to work on more complicated %layout
%\kun{\sout{shape}layout}
%of 
data structures such as graphs.
Graphs are essential in representing real-world relational information in social networks and e-commerce.
The capability to build high-quality recommender systems on graphs is indispensable to multiple businesses.
In these graph data structures, the data which we need to access is often not coalesced together, but scattered in memory (Figure~\ref{fig:dense_vs_sparse} (b)).
%
%A good example of such data structure is a graph.
%
%In recent years, there has been increasing interest of applying machine learning on graph-structured data.
%

%

\begin{figure}[t]
  \centering
  \includegraphics[width=\linewidth]{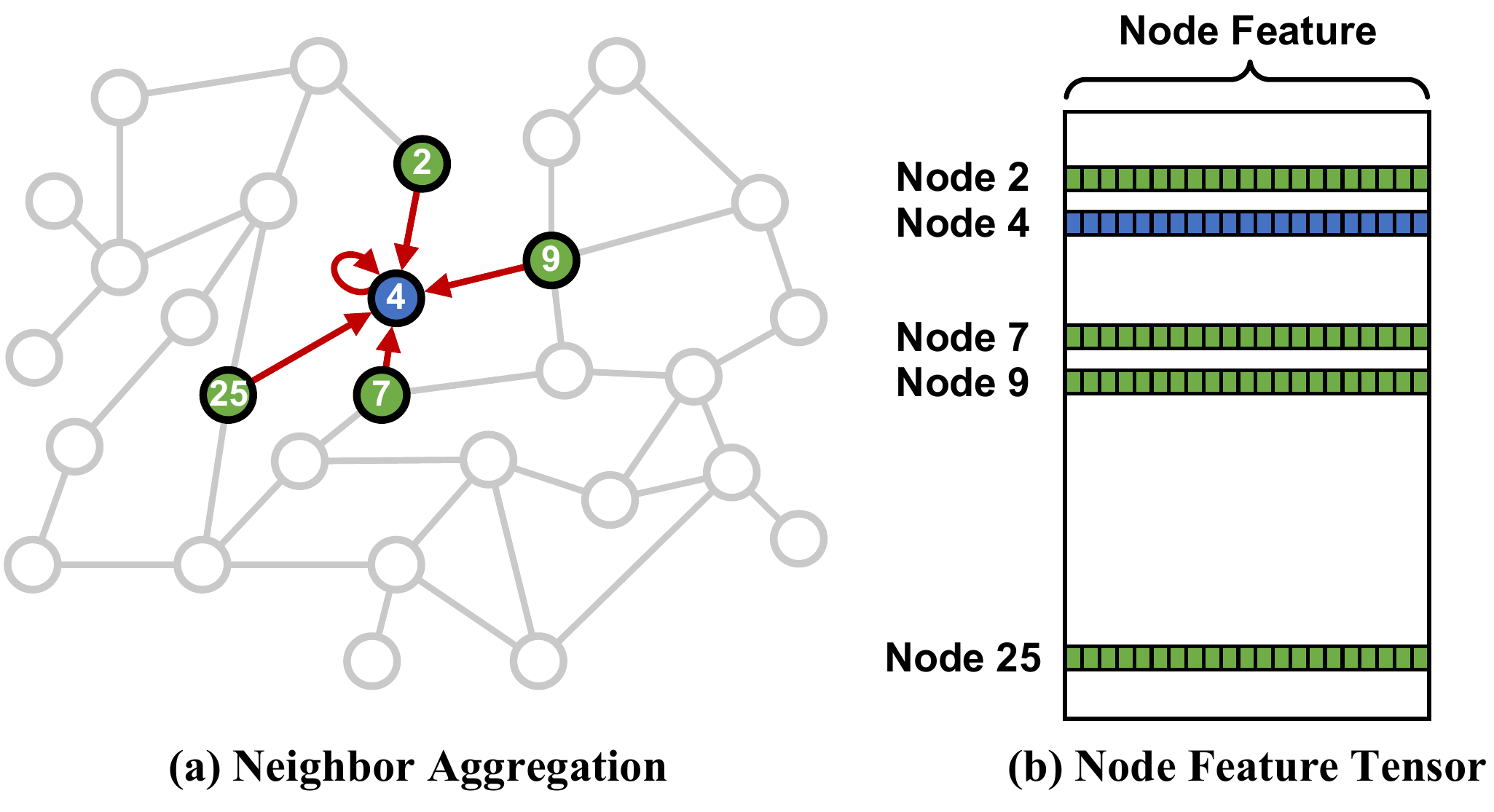}
  \caption{(a) A simple example of GCN training on single node. (b) An illustration of node features in memory. The neighboring nodes' features are scattered in memory.}
  \label{fig:background}
\end{figure}

One of the most successful adaptations of deep neural network models to graph data is Graph Convolutional Network (GCN)~\cite{kipf2017semi}.
The core idea of GCN is to create node embeddings by iteratively aggregating neighboring nodes' attributes
%\kun{Is ``attributes'' or ``latent vectors'' a better word/phrase here?}
using neural networks.
Due to its neighboring node's attribute lookup, training GCN requires accessing multiple scattered locations in memory.
In Figure~\ref{fig:background} (a), we show a simple example of GCN training.
To generate the embedding of node 4, we traverse the input graph and aggregate node 4's features alongside the features of all neighboring nodes in the node feature tensor.
%we traverse the input graph and aggregate the node features of all neighboring nodes and itself located in the node feature tensor.
%array \kun{Is ``tensor'' a  better word?}.
%
The example that we show here is only a toy example.
In real-world graphs, each node can be connected to thousands of nodes.
To collect relational information from those neighboring nodes, we may need to access thousands of scattered locations in memory.
Without a doubt, such data access patterns make the traditional block data transfer method %sub-optimal and 
ineffective.

\begin{figure}[t]
  \centering
  \includegraphics[width=0.9\linewidth]{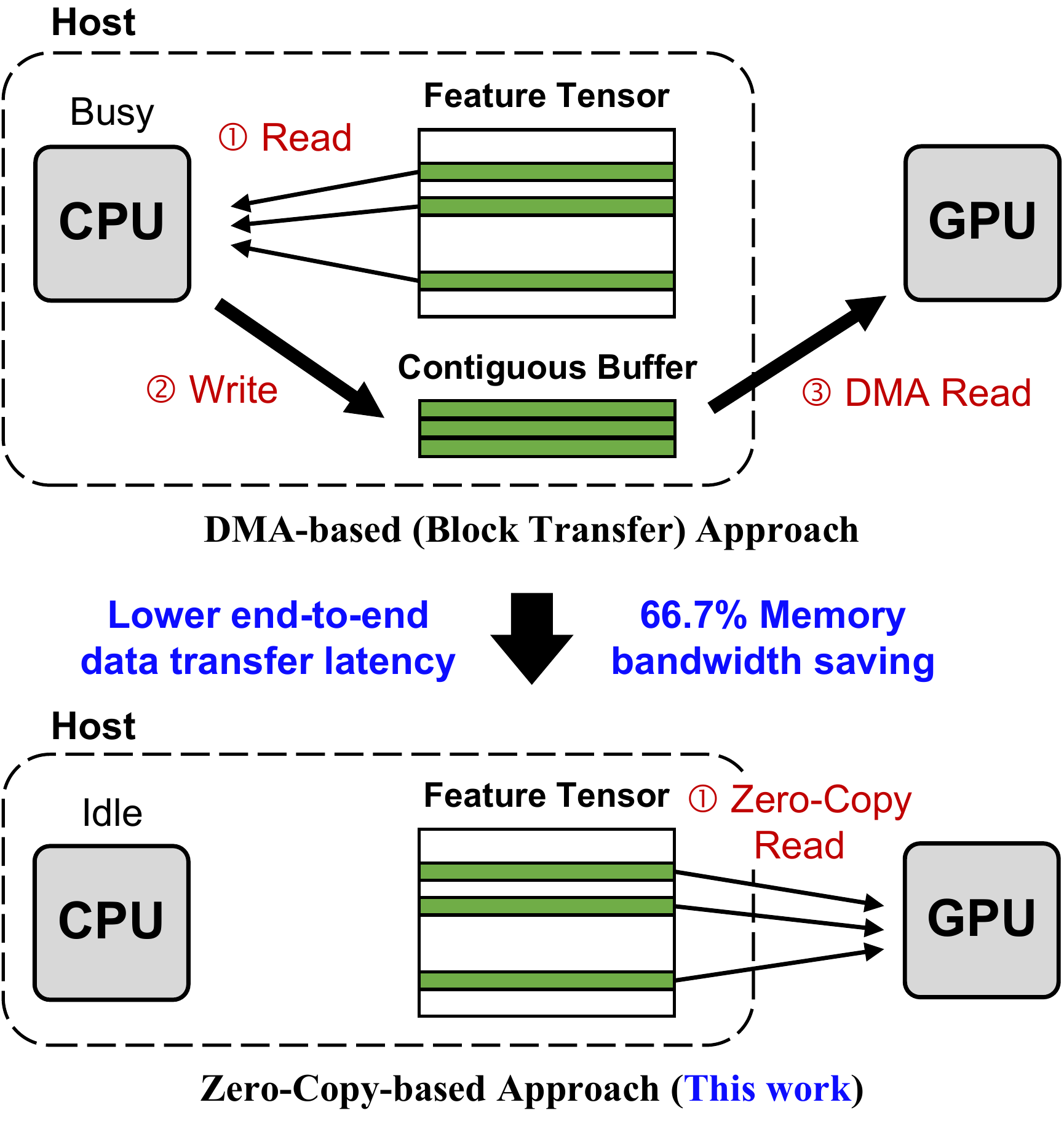}
  \caption{Workload comparison between DMA-based method and the proposed zero-copy-based method.}
  \label{fig:combined}
\end{figure}

%Therefore, i
In this work, we propose a processor-oriented, software-defined data communication architecture.
Instead of using DMA engines, we program GPU cores to directly access host memory with \textit{zero-copy memory access}.
%
%With uncomplicated DMA engines, we cannot efficiently access complicated data structures.
This approach allows the application developers to direct the GPU cores to exactly the locations that hold the data needed for computation.
Conventional wisdom may still argue that since the node feature data is in host memory, CPU has significant bandwidth advantage over GPUs and therefore DMA should be a better option because CPU can quickly gather the sparse features on the fly.
%consider increasing the efficiency of DMA by locally gathering node features with CPU.
However, recent work has shown that the ability to issue a massive number of concurrent memory accesses enables GPUs to tolerate latency effectively when accessing complicated data structures like graphs that reside in host memory~\cite{min2020emogi}.
%However, recent work has shown that GPUs with their ability to issue a massive number of concurrent memory accesses to tolerate latency can be effective in accessing data with complicated structures like graphs that reside in the host memory~\cite{min2020emogi}.
%
Therefore, in GCN training, if GPUs can make targeted fine-grain host memory accesses for sparse features while fully utilizing system interconnect (e.g., PCIe) bandwidth, the proposed approach can offer significant advantage over the DMA approach.
The removal of CPU gathering stage not only shortens data access latency for GPUs, but also greatly reduces the CPU and host memory utilization (Figure~\ref{fig:combined}).
%If successful, having the GPUs perform host memory access during GCN training eliminates the need to perform CPU gathering, which is wasting CPU cycles and host memory bandwidth. 
%\dc{this seems not a strong argument of the proposed idea. Would be a stronger claim to provide evidence that GPU performing host memory access directly would beat the alternative CPU-gathering-data option for GCN training.}
%
Offloading CPU workloads to GPUs also helps on training GCN with multiple GPUs as we can prevent the CPU becoming the bottleneck with increasing number of workers.

In order to propose the GPU-oriented data communication architecture for GCN training, we address three major questions in this work.
First, can zero-copy memory access fully utilize PCIe bandwidth while training GCN considering the long latency for accessing host memory?
Second, what would be the price of consuming GPU cores for zero-copy memory access?
Finally, after resolving the above two questions, can we show real end-to-end application performance benefit from our method?

In this work, we answer all three questions.
First, to maintain the best possible PCIe packet efficiency with zero-copy memory access, we propose an automatic data access alignment optimization in GPU data indexing kernel.
With our optimization, zero-copy PCIe bandwidth can match up to 93\% of block transfer PCIe bandwidth.
Second, we propose a novel CUDA multi-process service (MPS)~\cite{mpsguide} based resource provisioning optimization to minimize GPU resource consumption of zero-copy memory accesses.
Based on careful investigation of PCIe protocol and GPU architecture, we conclude that we can saturate PCIe even if only a few number of GPU cores are generating zero-copy accesses.
%\kun{Do we need to constrain this sentence with a specific GPU microarchitecture/model?}
%
Therefore, our optimization isolates only small portion of GPU resources for the zero-copy accesses and leaves the rest for computationally intense workloads.

Finally, we build an end-to-end zero-copy GCN training flow in PyTorch.
To enable zero-copy memory access, we devise a new class of tensor called "unified tensor".
This tensor provides an address mapping of host memory for GPUs so they can directly access host memory with zero-copy accesses.
By simply declaring multiple unified tensor instances for multiple GPUs, our GCN training flow can also support zero-copy access in multi-GPU training environment.
Our modifications are seamlessly integrated with the existing PyTorch framework and therefore we can quickly apply our method on existing GCN training applications.
We evaluate our design on multiple large graph datasets where the largest one has 111 millions of nodes and 1.6 billions of edges.
In a single-GPU training environment, our method is 16-44\% faster than the DMA-based method, but in a multi-GPU training environment, our method becomes 65-92\% faster than the DMA-based method.
Our method is efficient in hiding the remote sparse feature access time with the training time and can even match with the all-in-GPU-memory method for some graphs that fit in the GPU memory.
%\dc{Great. This DMA solution is done through gathering sparse feature on the fly, right? If so, some analysis/insights can be provided in the paragraph on top of this page (see my other comment there) to motivate why such a gain can be obtained. For example, is it true that the CPU needs to spend 65-92\% of the overhead to gather the data on the fly? If so, why so?}
%

In summary, the main contributions of this paper are as follows:
\begin{itemize}
    \item As opposed to the traditional DMA-based data communication architecture, we propose GPU-oriented, software-defined data communication architecture with zero-copy memory accesses for efficient sparse accesses to graph node features in GCN training.
    \item To improve the efficiency of zero-copy memory access, we propose automatic data alignment and a novel CUDA MPS based resource provisioning optimizations.
    \item We seamlessly integrate our modifications with the existing PyTorch framework for easier programming and show 65-92\% of end-to-end training performance gain. %\dc{Can add a bullet point to emphasize the performance gain through this new approach and also mention that this new framework together with libraries will be open-sourced to benefit the entire community.}
\end{itemize}

The rest of the paper is organized into the following sections. Section~\ref{sec:background} provides the necessary background for the proposed approach. Section~\ref{sec:impl} gives a brief overview of the proposed approach. Section~\ref{sec:eval} presents an experimental evaluation of the proposed approach. Section~\ref{sec:discuss} discusses potential future work. Section~\ref{sec:related} presents related works. Section~\ref{sec:conclusion} offers concluding remarks.
\section{Background}
\label{sec:background}
%\david{This section needs some work}

%This section presents the background of this work. We first introduces graph convolutional network and its neighborhood sampling approach. Then we introduce different ways to transfer data between CPU and GPUs, and discuss their advantages and shortcomings.
\subsection{Graph Convolutional Network}
The idea of Graph convolutional networks (GCN) \cite{lecunGCN, hamilton2017inductive, ying2019pinsage, GCNPierre, kipf2017semi, kipf2016variational, pmlr-v48-niepert16} started by an attempt to apply filters similar to convolutional neural networks (CNNs) \cite{CNN0} on graph structures.
Bruna et al. \cite{lecunGCN} was the first to propose the GCN model, where the authors utilized Laplacian filters as hidden layers to exploit global structure of the graph.
Such spectral construction is later adopted by many GCNs, including \cite{kipf2017semi, GCNPierre}.

GCNs are widely adopted in graph representation learning \cite{HamiltonYL17}, where GCN is trained to produce high-quality embeddings of the given nodes.
These embeddings can be used for
%the unstructured high-dimensional information in the graph to low-dimensional representation and paves the way for conveniently
performing several tasks such as link prediction and node classification.
Traditional representation learning algorithms, including node2vec \cite{node2vec} and DeepWalk \cite{DeepWalk}, are inherently shallow, transductive, and do not share parameters or utilize node attributes to encode node \cite{HamiltonYL17}. % \dc{why these are problems?} \kun{Added a one-liner to explain on this argument.}
These limits the representation power of the model, and disables the model to infer the representation when the nodes or edges are unseen in training.
GCN opens up the potential to develop algorithms to tackle these problems \cite{hamilton2017inductive, ying2019pinsage}. 

One severe issue with the early GCN is that the Laplacian filters in each layer are matrices whose dimension increases as the number of nodes in graph increases.
This effectively throttles the depth of GCN and the size of graph it can be applied due to the large memory footprint.
As an example, Kipf et al. \cite{kipf2017semi} presents a model for semi-supervised node classification using GCN. The simplified form of its forward-propagation function can be written as:
\begin{equation*}
H^{(l+1)}=f\left(H^{(l)},A\right)=\text{softmax}\left(AH^{(l)}W^{(l)}\right)
%    Z=f(X,A)=\text{softmax}\left(\hat{A}\text{ReLU}\left(\hat{A}XW^{(0)}\right)W^{(1)}\right)
\end{equation*}
where $A$ is an $N\times N$ adjacency matrix ($N$ is a number of nodes) representing the node connectivity, $H$ is an embedding table, $W$ is a weight table, and $l$ is a layer number.
$H^{(0)}$ is the input node feature table.
%where $\hat{A}=\Tilde{D}^{-\frac{1}{2}}\Tilde{A}\Tilde{D}^{-\frac{1}{2}}$ is of the same size as the adjacency matrix $A$, obtained through preprocessing: $\Tilde{A}=A+I_N$ and $\Tilde{D}_{ii}=\sum_j\Tilde{A}_{ij}$. %\dc{please double check to make sure this equation is right.} \kun{Yes this is correct. This basically means the degree of i is the sum of all connections of i with its immediate neighbours.}
%
%$W^{(0)}$ and $W^{(1)}$ are parameters of hidden layers and $X$ denotes input node attributes. 
Here, we can see the memory requirement of the operation is directly related to the size of $N$.

\subsection{Neighborhood Sampling}
\label{sec:back.sampling}
%\kun{Is this subsection too long? I feel hard to reduce it more.}
%\david{Can you reduce this section to at most two paragraphs as well. It doesn't need to be this detail.}\kun{shortened}

%\begin{figure}[t]
%  \centering
%  \includegraphics[width=\linewidth]{figures/neighbor_sampling2.pdf}
%  \caption{An example of the neighbor sampling of node pair (5, B) in training node representation. Each level indicates the nodes to aggregate for the connected node in upper level, i.e., the next aggregation layer. The node itself and its neighbors are distinct by different colors.}
%  \label{fig:neighbor_sampling}
%\end{figure}

To tackle the limitation of GCN, GraphSAGE~\cite{hamilton2017inductive} introduces neighborhood sampling and aggregating approach. %, able to generalize GCN algorithms.
By sampling fixed number of neighboring nodes instead of demanding the whole adjacency matrix, neighborhood sampling essentially reduces the computation and memory footprints and enables a fixed-size minibatching in both training and inference.
%
%With this approach, inductive learning can be also achieved: the representation of unseen nodes during training can be produced by the model as long as the nodes attributes are known because the hidden layers do not require the adjacency matrix of the graph any more. 

%
%Quoted from GraphSAGE \cite{kipf2017semi}, the forward-propagation function of the $k$-th layer is defined as follows.

%\begin{align}
%h_{\mathcal{N}_k(v)}^k&=\text{AGGREGATE}_k\left(h_u^{k-1},\forall u\in\mathcal{N}_k(v)\right)\label{equ:graphsage1}\\
%h_v^k&=\sigma\left(W^k\cdot \text{CONCAT}\left(h_v^{k-1},h^k_{\mathcal{N}_k(v)}\right)\right)\label{equ:graphsage2}
%\end{align}

%where $\mathcal{N}_k(v)$, $h^k_{\mathcal{N}_k(v)}$, and $W^k$ are the neighbor samples multiset of node $v$, the $k$-th layer output vector of $\mathcal{N}_k(v)$ and the $k$-th layer output vector of node $v$, respectively.
%
%$\text{CONCAT}(\cdot)$ and $\sigma(\cdot)$ are vector concatenation operation and activation function, respectively.
%
%$\text{AGGREGATE}_k(\cdot, \cdot)$ is the $k$-th layer aggregation function, which can be specified as mean, LSTM, etc.

GraphSAGE models are a sequence of aggregation layers, which can be LSTM, pooling, or mean operations.
The neighborhood sampling is applied to every neighboring node in every aggregation step.
GraphSAGE uses uniformly random selection process to sample the neighboring nodes, but other works such as FastGCN~\cite{chen2018fastgcn}, VR-GCN~\cite{pmlr-v80-chen18p} use more complex algorithms to determine the neighboring nodes that need to be sampled.
The commonly used hyperparameters for the neighborhood sampling size $S_{layer}$ are $(S_1, S_2) = (10, 25)$ and $(S_1, S_2, S_3) = (10, 10, 10)$.
It is uncommon to go beyond the three layers of sampling due to the exponential growth on the number of nodes that need to be sampled.
After the sampling, a sub-graph which only contains the sampled nodes is created so the computation kernel knows how to aggregate the node features of interest.
Over different epochs of training, a new sampling is done to increase the learning entropy and to cover more corner cases.
The exact implementation of the sampling process is framework-dependant.
In case of deep graph library (DGL)~\cite{wang2019dgl}, this part is written in C++ with OpenMP to maximize the performance, but PyTorch-Geometric~\cite{Fey/Lenssen/2019} simply uses a python code.

If the entire node feature table is not fitting in the GPU memory, the sampled nodes' features must be transferred after each sampling step~\cite{ying2019pinsage}.
Since the sampled nodes' features are scattered over the feature table, the current GCN implementations in PyTorch or TensorFlow, the frameworks which use DMA as a data transfer method, require the features to be collected into a dense format prior to the data transfer.

\subsection{CPU--GPU Data Communication}
%\kun{Not to mention implicit sync here since the potential implicit call in explicit copy API calls seems too detailed and not directly connected to the serialization mentioned in evaluation.}
%\kun{Last paragraph on the data pattern optimizations in need in order to achieve good performance when using zero-copy access is completely removed.}

CUDA provides developers with three ways to transfer data between host and GPUs: (1) DMA APIs, (2) automatic page migration, and (3) zero-copy access \cite{NvidiaUVA}.

As the first method, CUDA provides both synchronous and asynchronous APIs to copy data among host and devices.
The two most commonly used APIs are \texttt{cudaMemcpy()} and \texttt{cudaMemcpyAsync()}.
Both functions take a source pointer, a destination pointer, a data size, and a data transfer direction.
%
%There are also other APIs to provide functionality, such as multi-dimensional data copy and prefetching.
%
For this method, DMA engine is used for the data transfer.
DMA engine is efficient in transferring a single large data block, but sub-optimal for transferring small sized data due to the DMA request setup latency caused by user program <-> operating system interactions.
According to Pearson et al.~\cite{pearson19}, to make the effective bandwidth of DMA to about 90\% of maximum PCIe 3.0 x16 bandwidth, the data block size should be at least 256KB.
With 64KB of data block transfer, the DMA efficiency drops to less than 50\% of the maximum PCIe 3.0 x16 bandwidth.
%if the source data is pinned; otherwise, a copy will be performed from the source to a pinned buffer before the DMA transfer is initiated, which occupies the host CPU and reduces the effective memory bandwidth \cite{CUDACommChar}.

Page migration is the second way.
To provide convenience to programmers, NVIDIA introduced the Unified Virtual Memory (UVM) \cite{P100Whitepaper,V100Whitepaper,A100Whitepaper, UVMPrimer, pearson19}.
%to systems accelerated by Nvidia GPUs
%via changes and new components in hardware and drivers \cite{UVMPrimer}.
%
Data pointers to the memory regions managed by the UVM driver can be dereferenced by both GPU kernels and CPU functions.
When a processor (either CPU or GPU) attempts to access a page that it does not own in its local memory, the accessed page needs to be migrated from a remote location.
%When the memory region is accessed by a remote processor, which can be either CPU or GPU, the accessed page will be migrated to the local memory of the remote processor before the execution continues. \kun{Done rephrease.}
%When the pointers are accessed by another processor, whether CPU or GPU, its current owner for the first time, the corresponding page will be migrated to the processor before the execution is resumed\david{Can you rephrase this sentence}.
%
Similarly, if other processor accesses this page later on, page migration to that processor will be triggered.
%
%Since Pascal microarchitecture, on GPUs page migrations are triggered by page faults when the missing page is accessed for the first time \cite{UVMPrimer}.
%
The minimum migration granularity is identical to the system page size (4KB), but it can be as large as 2MB.
%
%may vary but the minimal is 4KB \cite{min2020emogi}.
%
UVM makes programming easier by removing the need of explicit call of \texttt{cudaMemcpy()} by users.
%, especially in cases where deep copies are needed.
%
To allocate the UVM-backed memory region, programmers simply need to call \texttt{cudaMallocManaged()} with the desired size.
%To enable UVM in a memory region, programmers only need to allocate it using \texttt{cudaMallocManaged()} API.
However, the programmer-friendly page migration is not designed to be a performant mechanism of data transfer.
Its performance is limited with irregular access patterns due to high page miss rate.
%
%The first issue is the inherent poor spatial data locality in graph-structured data.
%
%For example, in each minibatch in training the node representation, nodes to access are either in pairs making up the batch or as neighbors of the nodes in the pairs; therefore, these nodes attributes are likely to scatter on many pages, instead of taking up most of the portion in a few pages.
%
This leads to an excessive amount of page faults that stall the execution and create I/O read amplification.
With larger discrepancy between the dataset size and the GPU memory size, there will be more frequent page migrations incurred by severe page thrashing.

Finally, CUDA enables zero-copy access, which is also known as direct access.
In a zero-copy access, GPU sends a cacheline-sized memory request directly through external interconnect (e.g. PCIe), without explicit data copy or page migration that will happen in the aforementioned two methods.
The source memory region can be the host memory, peer PCIe devices, or other GPUs connected over NVLink.
Zero-copy is useful in accessing fine-grained data, but it needs GPU cores to be engaged in generating memory requests.

%There might be misbelief that zero-copy data transfer is slower than DMA-based approach.
%
%Actually, as Min et al. \cite{min2020emogi} pointed out, after carefully performing data transfer pattern optimization, specifically merging requests and aligning the base address to the request size, the performance will get close to the bus bandwidth.
%
%In this work, we introduce circular shift stage to the indexing kernel to issue aligned 128 bytes PCIe requests.
%
%This is the maximal PCIe payload size in this scenario, where each threads in a warp of 32 threads accesses a 4 bytes single precision floating point.
%
%See Section \ref{sec:impl_aligned} for details.

%The memory region eligible for zero-copy access needs to be pinned and memory-mapped to an address on the shared bus \cite{min2020emogi}.
%
%There are three APIs combinations to achieve that \cite{min2020emogi}.
%
%In this work, we choose to apply \texttt{cudaHostRegister()} to register a regularly-allocated host memory region and use \texttt{cudaGetDevicePointer()} to retrieve the device pointer.

\section{GPU-Oriented Data Communication Architecture}
\label{sec:impl}
Due to the wide spread use of DMA-based data communication architecture, there are some number of system-level modifications that must to be established to support our GPU-oriented data communication architecture in the higher-level programming models.
In this section, we first describe how we enable zero-copy accesses in PyTorch and then we discuss some of the technical aspects of zero-copy access to identify its weaknesses and how to overcome them.
Finally, we describe the end-to-end GCN training flow using zero-copy accesses.

\subsection{Zero-Copy Enablement in PyTorch}

For GCN training, we use PyTorch which is one of the most popular python-based ML frameworks.
However, including PyTorch, there are no python-based ML libraries which naturally support zero-copy access for GPUs.
To overcome such issue, we create an extension of the existing PyTorch implementation with several modifications in its source code.

\begin{table}[t]
  \caption{PyTorch Tensor Class Comparison.}
  \label{tab:tensor}
  \begin{tabular}{cccc}
    \toprule
         & \multicolumn{2}{c}{Existing} & \textbf{This Work} \\
    \cmidrule(lr){2-3}\cmidrule(lr){4-4}
        Context & CPU & CUDA & Unified \\
        Worker & CPU & GPU & GPU \\
        Data Storage & Host Memory & GPU Memory & Host Memory\\
    \bottomrule
  \end{tabular}
\end{table}

In PyTorch, data is allocated through a class called "tensor".
The physical location of data is determined by a context value which is passed to the class upon a declaration (Table~\ref{tab:tensor}).
In the current implementation of PyTorch, processing units can only compute data located within their own local memories.
For example, to perform a GPU-accelerated matrix multiplication on CPU tensor, a new tensor with CUDA context should be created.
When the new CUDA tensor is created based on the old CPU tensor, PyTorch automatically calls DMA to copy the data in the host memory to the GPU memory.

In our design, we aim to aggressively avoid the implicit DMA data copy performed by PyTorch.
We give GPUs direct access to tensor data in the host memory by mapping the host-memory data pointers into the GPU address space.
To achieve our goal, we create a new class of tensor with a new "unified" context.
A tensor with this new context can be declared from any existing CPU tensors.
Upon the declaration, the tensor calls the \texttt{cudaHostReigster()} and \texttt{cudaHostGetDevicePointer()} CUDA APIs internally.

Calling \texttt{cudaHostReigster()} page-locks the given CPU tensor data and \texttt{cudaHostGetDevicePointer()} maps page-locked data into the GPU address space
%
%After the memory-mapping, \texttt{cudaHostGetDevicePointer()} 
and returns a device pointer which can be used in GPU kernels for zero-copy accesses.
There are several other ways of allocating a host memory space for zero-copy such as  \texttt{cudaMallocHost()}, \texttt{cudaHostAlloc()}, or \texttt{cudaMallocManaged()} with \texttt{cudaMemAdvise()}, but these methods have some limitation for multi-GPU training, which we will explain in Section~\ref{sec:impl.multi}.

\begin{lstlisting}[float,caption={PyTorch Programming with Unified Tensor},label={lst:unified_tensor}]
import torch

<@\textcolor{blue}{\# Input tensor data in host memory}@>
input_tensor = torch.randn([100], device="cpu")

<@\textcolor{blue}{\# CUDA tensor created, data copied by DMA (e.g. cudaMemcpy())}@>
gpu_tensor = input_tensor.to(device="cuda")

<@\textcolor{blue}{\# Unified tensor created, no data copy occurs}@>
unified_tensor = input_tensor.to(device="unified")

<@\textcolor{blue}{\# gpu\_tensor data comes from GPU memory}@>
<@\textcolor{blue}{\# unified\_tensor accessed through zero-copy access to host memory}@>
<@\textcolor{blue}{\# Computation done by GPU}@>
output = gpu_tensor + unified_tensor
\end{lstlisting}

Besides the pointer manipulation, other existing PyTorch tensor mechanisms remain the same and therefore there are no noticeable functional differences introduced to the end-users.
Listing~\ref{lst:unified_tensor} shows a simple vector addition example in PyTorch using unified tensor.
From the code, we can see declaring the unified tensor is as simple as declaring the existing CUDA tensor.
While the CUDA tensor is created by explicitly copying data from the CPU tensor, the unified tensor only creates a mapping to the host memory for the GPU.
We have empirically measured that the GPU memory usage by the memory mapping is about 1/512 of the data size.
Therefore, while the CUDA tensor will immediately fail on declaration if the data size is larger than the GPU memory capacity, the unified tensor can hold up to 512 times more.

\subsection{Improving Zero-Copy Efficiency Over PCIe}
One of the common misconceptions of zero-copy access is its low data transfer efficiency compared to the DMA-based methods~\cite{gangulyadaptive}.
The misconception is mainly coming from the fact that the users are treating the zero-copy without any specific care.
However, as the zero-copy access requests are made over PCIe, it is important to understand how the zero-copy accesses interact with PCIe.
In this section, we take a deep-dive into the technical aspect of PCIe protocol and its interaction with GPUs. We then present two important techniques for maximizing the zero-copy efficiency during GCN training.

\subsubsection{Aligned Memory Access}
\label{sec:impl_aligned}
Even though our purpose of using zero-copy is to make fine-grained memory accesses to the host memory, it is still desirable to make coarser-grained PCIe memory requests whenever possible for a couple of reasons.
First, each PCIe packet has 12--16 bytes of header overhead.
Therefore, to compensate the overhead, it is better to increase the payload size by requesting a larger memory request.
Second, PCIe devices have a hard limit on the number of outstanding requests they can create.
Since the PCIe round trip time (RTT) is very long (1--5us, variable), it is necessary to submit multiple read requests in a pipelined fashion to fully occupy the interconnect.
However, if we squander the capacity by generating too many small read requests, it becomes difficult to fully tolerate the latency and utilize the PCIe bandwidth.
The numbers of maximum outstanding read requests for PCIe 3.0 and PCIe 4.0 are 256 and 768, respectively.

\begin{figure}[t]
  \centering
  \includegraphics[width=\linewidth]{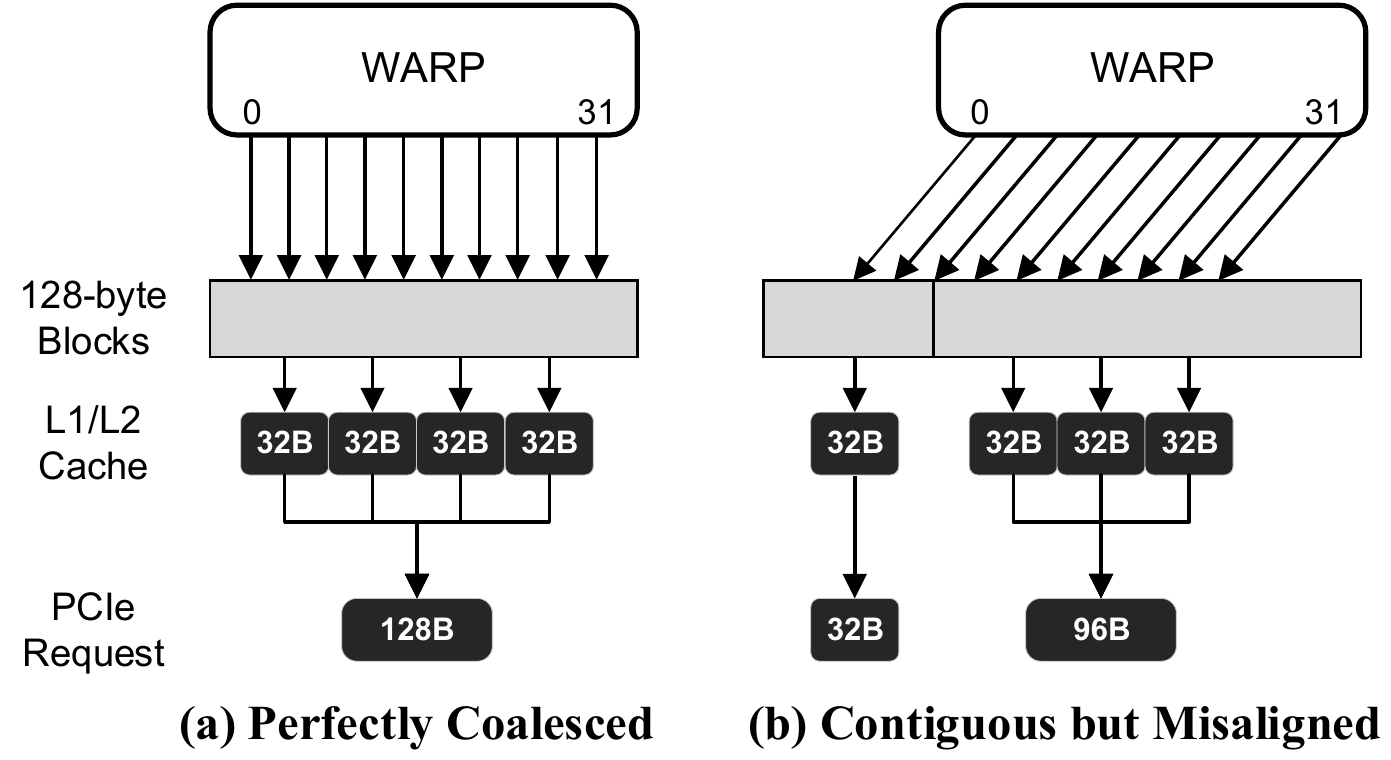}
  \caption{(a) A perfectly coalesced 128-byte access from a warp. (b) A warp accessing a misaligned data needs to generate multiple PCIe requests.}
  \label{fig:misalign}
\end{figure}

\begin{figure}[t]
  \centering
  \includegraphics[width=\linewidth]{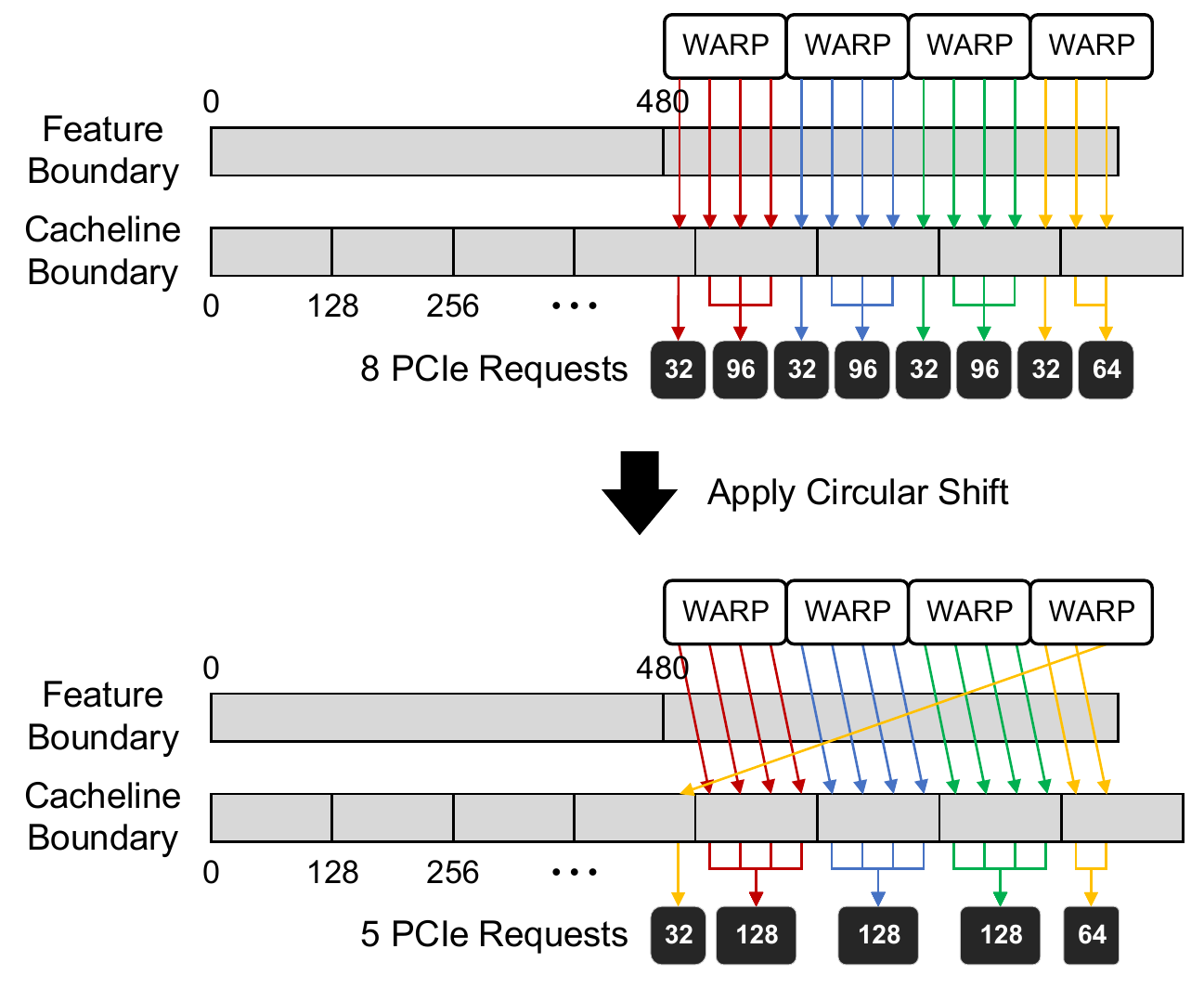}
  \caption{Circular shift optimization explained. Circular shift transforms memory requests into a GPU cacheline-friendly way.}
  \label{fig:auto_alignment}
\end{figure}

Now, with all that in mind, how do we generate coarser-grained PCIe requests?
According to Min et al.~\cite{min2020emogi}, to make PCIe read requests more efficient, the same technique used for the GPU memory coalescing~\cite{howtocoalesce} can be used.
In Figure~\ref{fig:misalign}, we explain two cases where (a) memory accesses from a warp are contiguous and aligned with the GPU cacheline, and (b) memory accesses from a warp are contiguous but misaligned with the GPU cacheline.
In case of (a), the accesses from the threads in a warp are perfectly coalesced and the coalesced requests becomes a single 128B PCIe read request.
In case of (b), the accesses from a warp are scattered over two GPU cachelines and they result in generating two separate PCIe read requests.
The possible memory access granularities are 32B, 64B, 96B, and 128B, while 32B is a single sector size of GPU cacheline~\cite{profileguide}.
Each GPU cacheline is composed of four sectors.

\begin{lstlisting}[float,caption={GPU Indexing Kernel and Automatic Alignement},label={lst:aligned}]
#define WARP_SIZE 32

__global__ void index(float* dst, float* src,
                      int* idx_list, int feat_size,
                      int numElem) {
  int linearIdx = blockDim.x * blockIdx.x
                  + threadIdx.x;
  
  for (int i = linearIdx; i < numElem;
       i += blockDim.x * gridDim.x) {
    int dstIdx = i / feat_size;
    int offset = i % feat_size;       
      
    int dstStart = dstIdx * feat_size;
    int srcStart = idx_list[dstIdx] * feat_size;
    
    int dstOffset = offset + dstStart;
    int srcOffset = offset + srcStart;
    
    <@\textcolor{forestgreen(web)}{// Cacheline-size-aware circular shift stage added}@>
    <@\textcolor{blue}{if (feat\_size > WARP\_SIZE \&\& feat\_size \% WARP\_SIZE) \{}@>
      <@\textcolor{blue}{int diff = (dstStart - srcStart) \% WARP\_SIZE;}@>
      <@\textcolor{blue}{diff = diff < 0 ? diff + WARP\_SIZE : diff;}@>
    
      <@\textcolor{blue}{dstOffset += diff;}@>
      <@\textcolor{blue}{srcOffset += diff;}@>
    
      <@\textcolor{blue}{if (srcOffset < srcStart) \{}@>
        <@\textcolor{blue}{dstOffset += feat\_size;}@>
        <@\textcolor{blue}{srcOffset += feat\_size;}@>
      <@\textcolor{blue}{\}}@>
      <@\textcolor{blue}{else if (srcOffset >= srcStart + feat\_size) \{}@>
        <@\textcolor{blue}{dstOffset -= feat\_size;}@>
        <@\textcolor{blue}{srcOffset -= feat\_size;}@>
      <@\textcolor{blue}{\}}@>
    <@\textcolor{blue}{\}}@>
    
    dst[dstOffset] = src[srcOffset];
  }
}
\end{lstlisting}

Of course, we would not need to worry about the misaligned accesses if the node feature objects always start at 128B boundaries and the sizes of node features are always multiples of 128B, but it is very unlikely to be so in reality.
For example, if a certain dataset's node feature size is 480B, accessing the second node feature will start from accessing 480th byte in memory address.
In this case, we are off by 32B from the closest 128B boundary (512B).
To automatically resolve such issue, we add a circular shift stage in the PyTorch indexing CUDA kernel.
In Listing~\ref{lst:aligned}, we show the circular shift stage code we added, but in a simplified manner.
The shifting stage is aware of the GPU cacheline size and shifts the memory access indices by calculating the offset between the nearest 128B aligned location and the current indexing location.
The visualization of our circular shift mechanism is shown in Figure~\ref{fig:auto_alignment}.
In this example, we want to access the second node feature with zero-copy access where each node feature size is 480B.
Without the optimization, each warp start reading from misaligned locations and end up generating 8 PCIe requests.
However, once our optimization is applied, the warps adjust their indexing locations and try to generate aligned memory accesses as much as possible.
In this example, the total number of PCIe read requests is reduced to 5.

We do not apply the circular shift stage if the node feature size is less than the GPU cacheline size or if it is already a multiple of the GPU cacheline size.
All these adjustments are transparent to the high-level programmers as a result of our modifications to PyTorch source code.
%
%As such, there is no programmer effort required for solving the memory alignment problem. 

\subsubsection{Asynchronous Operations and Resource Provisioning}
\label{sec:impl.async}

One important distinction of our design is that zero-copy accesses are done by GPU kernels.
In other words, the other following GPU kernels need to wait until the zero-copy kernel is finished even if all it's doing is simply reading the host memory. %, we need to utilize GPU cores explicitly.
However, like in many other ML algorithms, GCN can also greatly benefit from overlapping data communication time and training time, which naturally happens in DMA-based methods.
To achieve the best training performance, we must devise a way to overlap the training GPU kernels and the zero-copy GPU kernels in our design.
%

%\begin{figure}[t]
%  \centering
%  \includegraphics[width=\linewidth]{figures/concurrency.pdf}
%  \caption{GPU resource distribution with zero-copy kernel.}
%  \label{fig:concurrency}
%\end{figure}

Normally, concurrency and overlapping activities can be accomplished by using CUDA streams.
CUDA streams allow GPU kernels and API service activities in different streams to execute in arbitrary order so as to enable overlapped operations.
Unfortunately, there are several situations where achieving the concurrency is impossible.
First, there are several blocking CUDA APIs such as \texttt{cudaMalloc()}, \texttt{cudaFree()}, and \texttt{cudaEventQuery()} that serialize the GPU operations.
In current implementation of PyTorch, some of the listed APIs are called in  background implicitly, such as by memory allocation manager.
If one of the CUDA APIs are called in between the zero-copy GPU kernel and the training GPU kernel, the latter GPU kernel must wait until the entire operation of the earlier GPU kernel to be finished.
Second, when the GPU resources are completely consumed by a current GPU kernel, the following kernel must wait until the resources are released.
In general, most of the GPU kernels try to occupy as much as of GPU resources they can and the serialization situation is very likely to occur.
%to finish their operations as soon as possible. \kun{This sentence seems hard to read.}
%

However, in fact, we have missed a fundamental question here.
Before we think about the concurrency, how much of GPU resource do we need for the zero-copy GPU kernels?
If we need the entire GPU resource to fully utilize the PCIe bandwidth, then there is no point of attempting to achieve the concurrency in the first place.
This is the core question which needs to be answered to verify the validity of idea of overlapping zero-copy and training GPU kernels.

To answer to this question, we explore the architecture details of NVIDIA GPUs.
In NVIDIA GPUs, to better utilize computation units and to hide long GPU memory access latency, each single physical core may have multiple active warps to issue instructions from~\cite{volkov2016understanding}.
In this way, the physical core won't be stalled when some of the warps are waiting for the completion of their memory requests.
%\kun{This sentence is rewritten. Please review the change by comparing it with the original in the comment.}
%In NVIDIA GPUs, to better utilize computation units and to hide long GPU memory access latency, multiple logical threads can be queued into a single physical core while waiting for their memory requests to be finished~\cite{volkov2016understanding}.
%
%Therefore, the number of physical CUDA cores we need to reserve for the zero-copy accesses is much smaller than the actual number of PCIe read requests we want to generate.
Therefore, the number of physical GPU cores we need to reserve is much smaller than the number of memory requests we want to generate.
%\kun{``in theory'' seems redundant.}
%

\begin{table}[t]
  \caption{NVIDIA RTX 3090 Specifications.}
  \label{tab:rtx3090}
  \begin{tabular}{lc}
    \toprule
        \multicolumn{1}{c}{Category} & \multicolumn{1}{c}{Specification} \\
    \midrule
        PCIe Generation & 4.0\\
    \midrule
        Max \# of Outstanding PCIe 4.0 Read Requests & 768\\
    \midrule
        \# of Multiprocessors & 82\\
    \midrule
        \# of Threads per Multiprocessor & 1536\\
    \midrule
        \# of Threads per Warp & 32\\        
    \bottomrule
  \end{tabular}
\end{table}

\begin{figure*}[t]
  \centering
  \includegraphics[width=\textwidth]{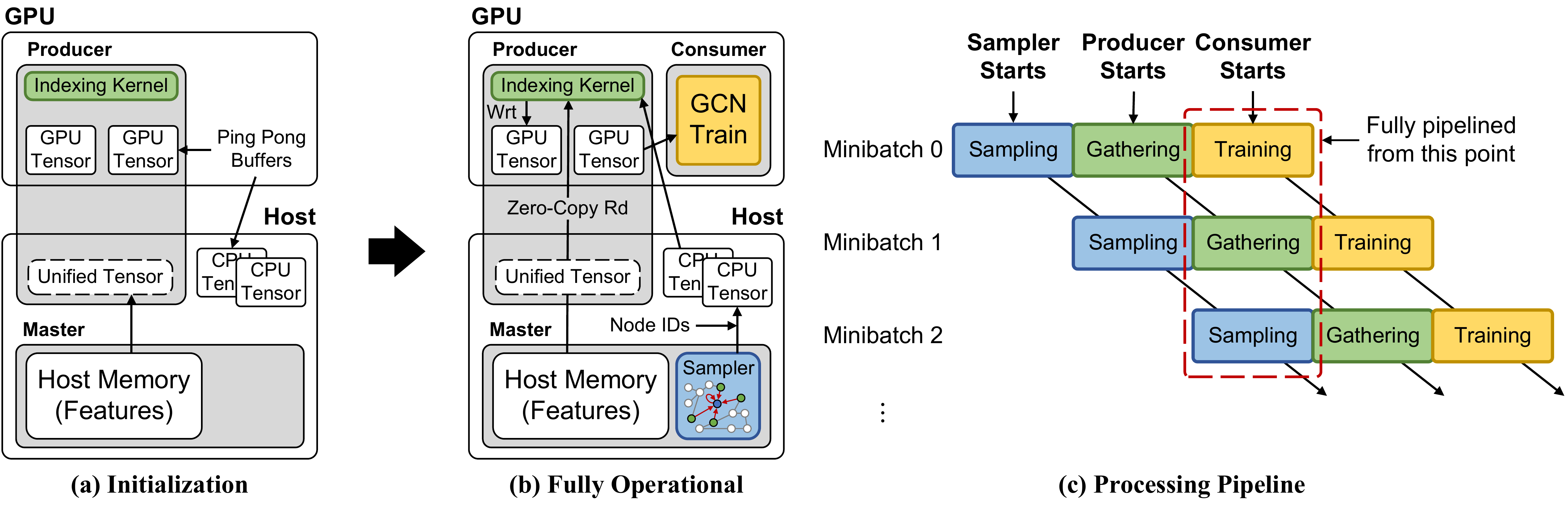}
  \caption{GCN training flow with zero-copy accesses. Only the operations related to data accesses are  shown. (a) We setup unified tensor and the returned pointer is passed to GPU for zero-copy accesses. (b) The sampler generates node IDs used by the producer and the producer gathers scattered node features in the host memory. The consumer uses the gathered node features for training. (c) A visualization of processing pipeline.} %\kun{According to the figure and the procedure to create the unified tensor here, the reviewer might challenge that the unified tensor is a view instead of a real materialized tensor.}\david{why this is something to be challenged?}\kun{Sorry I mistook. Please ignore this.}}
  \label{fig:singlegpu}
\end{figure*}

As we discussed in the previous Section~\ref{sec:impl_aligned}, there is a hard limit on the number of outstanding PCIe read requests that
%\kun{\sout{where} that}
PCIe devices can generate at a given moment.
Therefore, if we can prove that we only need small amount of GPU resources to fully saturate the limit, it is worthwhile to seek for a way to achieve the concurrency.
In Table~\ref{tab:rtx3090}, we list the specifications of NVIDIA RTX 3090 GPU which we use for our evaluations.
At any given moment, the GPU cannot generate more than 768 outstanding PCIe read requests.
To identify the portion of GPU resource we need to generate 768 outstanding PCIe read requests, we perform the following calculation.
First, lets assume each warp's memory requests are coalesced to a single PCIe read request, and lets ignore the payload size for now.
In this case, we need 768 warps available to the scheduler to reach the PCIe 4.0 limit.
Since each streaming multiprocessor (physical processor) can hold up to 1,536 threads at a given moment, each multiprocessor can sustain up to 1,536 / 32 = 48 outstanding PCIe read requests.
Now, we have 82 multiprocessors in RTX 3090, so the amount of GPU resource that we need to reserve for the zero-copy GPU kernel is about 16 / 82 = 19.5\%.
However, this is the upper bound for the extreme case.
If we assume we can always generate 128B PCIe read requests, we can saturate the PCIe 4.0 bandwidth with much fewer outstanding requests.
For example, the measured maximum PCIe bandwidth with \texttt{cudaMemcpy()} in RTX 3090 is 25.8GB/s and if we assume RTT (Round-Trip-Time) of PCIe is 1.5us~\cite{10.1145/3230543.3230560}, the number of outstanding requests that we need to sustain is (25.8GB/s) / (128B) $\times$ 1.5us = 324.6.
That is, assuming all PCIe requests are 128B in size, we need to reserve only 8.2\% of the total GPU resource for the zero-copy GPU kernel. In reality, since some of the requests will be smaller, this number is a lower bound and the actual number will be somewhat higher.
In short, even if we try to maximize the zero-copy GPU kernel efficiency, there is at least 80\% and up to 91.8\% of the GPU resources available for other workloads.

Now, finally, since we realized how much of GPU resource should be allocated for the zero-copy kernel, we explore the method to enforce the limitation in practice.
Fortunately, NVIDIA GPUs already provide support for limited execution resource provisioning through CUDA multiprocessing service (MPS)~\cite{mpsguide}.
MPS is originally designed to improve quality of service (QoS) between different clients' workloads, but we utilize this service to control the resource utilization of the zero-copy GPU kernel.
To assign different resource limitations to different kernels, the kernels must be running in different processes.
Since PyTorch already supports multiprocessing programming model, it is simple to launch the zero-copy GPU kernel and the training GPU kernel in two separate processes.
Before we launch the zero-copy GPU kernel, we modify the GPU resource limitation to $X$\% with the \texttt{nvidia-cuda-mps-control} utility.
Next, before we launch the training GPU kernel, we also modify the resource limitation to $(100-X)$\%.
In our PyTorch code, the whole process is scripted for an easier use.
It would be more elegant if the resource limitation can be configured in the user CUDA code instead of the MPS utility, but currently CUDA does not support such functionality.
Another side benefit of the multiprocessing approach is that the different GPU kernels running in different processes are not affected by the other processes' blocking CUDA API calls.
With our approach, zero-copy accesses can saturate the PCIe bandwidth while leaving majority of GPU resources opened for other computationally intensive workloads.

With this optimization, we can basically transform the GPU cores into an intelligent DMA engine which can asynchronously perform complex data accesses such as data dependant index calculations and fine-grained host memory accesses.
This optimization can be also useful for some of the workloads which utilize  peer-to-peer GPU memory accesses with zero-copy accesses.

\subsection{Workload Scheduling}
\label{sec:impl.scheduling}
In this section, we combine all the implementation details we discussed in the previous sections, and explain the overall flow of our GCN training with zero-copy accesses enabled.
%
%\kun{\sout{using zero-copy accesses} with zero-copy accesses enabled}.
%using zero-copy accesses.
%
In Figure~\ref{fig:singlegpu} (a), we show the initial tensor allocations during the initialization step.
First, we map the whole node feature tensor
%\kun{\sout{array} tensor}
into the GPU address space by using the unified tensor.
This unified tensor holds a memory pointer which GPU can use in its kernel to generate zero-copy accesses to the node feature tensor.
%\kun{\sout{array} tensor}.
Next, we create two sets of ping pong buffers for interprocess communications.
The goal of using ping pong buffers is to remove the usage of locking mechanisms between two different processes sharing data and to allow them to start working for the next minibatch immediately after finishing their current works.
In our design, each process needs to be synchronized just once per minibatch.

After the initialization, the training pipeline begins from the sampler process
%\kun{Is this a procedure or a process?}
randomly selecting nodes and collecting their neighbors' node indices (Figure~\ref{fig:singlegpu} (b)).
%
%The detailed process of sampling is explained in Section~\ref{sec:back.sampling}.
%
Once all the node indices are identified, the combined list is transferred to the producer process running on GPU for the zero-copy accesses.
The list of node indices is transferred over DMA as it is contiguous and small.
Once the node features are all gathered into one of the ping pong buffers, the producer notifies the consumer to train on the new minibatch data as soon as it is ready.
Since the GPU ping pong buffers are located in the same GPU memory, it naturally makes sense for the consumer to directly access the buffer owned by the producer instead of copying it to its own space.
To achieve this, we utilize CUDA interprocess communication (IPC) APIs.
With the CUDA IPC APIs, two different GPU kernels running on different processes can share the same GPU memory space without data copies in between.
This specific GPU pointer sharing procedure is implemented in the PyTorch \texttt{Queue} class and we utilize it for our application.
The ping pong buffers are statically located for the entire training process and therefore the pointer sharing needs to be done only once at the beginning of the producer process.

From the user's point of view, the training process is pipelined in a sampler \textrightarrow ~ producer \textrightarrow ~ consumer order (Figure~\ref{fig:singlegpu} (c)).
Except the unified tensor declaration, the rest of our end-to-end GCN training implementations is developed with the existing PyTorch functionalities, and this makes our method more accessible for the existing users.
Our modifications are isolated into the data transfer portion of the GCN training and the algorithm remains unaffected.

\begin{figure}[t]
  \centering
  \includegraphics[width=\linewidth]{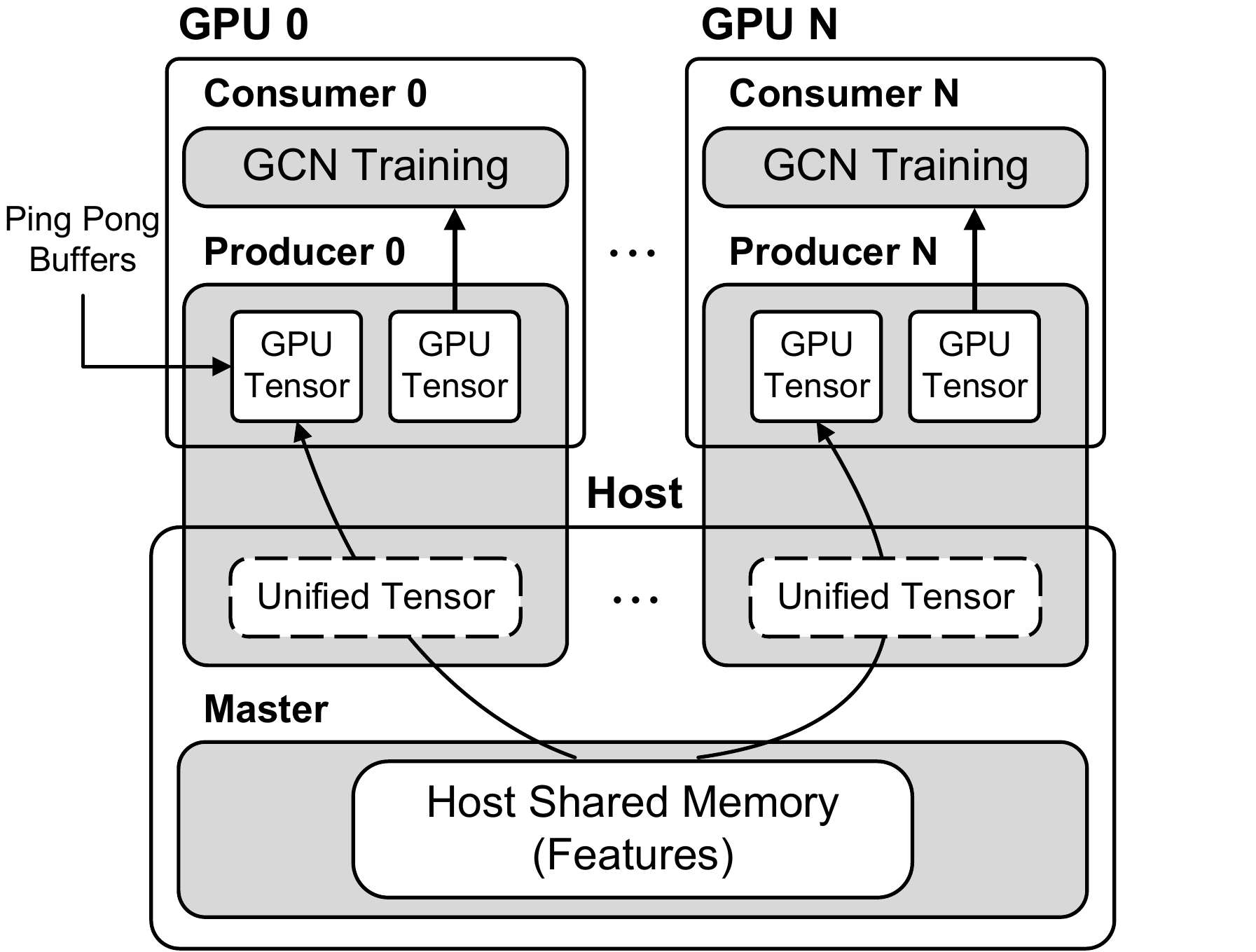}
  \caption{Simplified view of Multi-GPU GCN training flow. All unified tensors provide identical mappings. Sampling processes and indexing kernels are omitted in this diagram.}
  \label{fig:multigpu_simple}
\end{figure}

\subsection{Multi-GPU Training}
\label{sec:impl.multi}
The final challenge of our method is supporting multi-GPU training environment.
Multi-GPU training is one of the keystones of modern ML for reducing the training time, and the existing DMA-based method already supports the multi-GPU training.
Therefore, it is infeasible to propose a method which can only support a single-GPU training environment.

For multi-GPU training, we take the data parallelism approach used in DGL~\cite{wang2019dgl}.
In the original DGL implementation, the multi-GPU training is done by increasing the number of sampler-consumer pairs and assigning one GPU to each pair.
On top of the DGL implementation, we add the GPU-based producer process into each pair.
The DGL implementation does not have a dedicated producer process as it assumes the node feature data is collected by the sampler and transferred into each GPU's memory.
The simplified diagram of our multi-GPU training design is shown in Figure~\ref{fig:multigpu_simple}.

The main difficulty of multi-GPU training with zero-copy accesses lies on sharing the same host memory space across different GPUs running in different processes.
In general, sharing the host memory space across different GPUs is simple when the kernels are launched by a single process.
In this case, programmers simply need to allocate a memory space by either calling \texttt{cudaHostAlloc()} with a \texttt{cudaHostAllocPortable} flag or calling \texttt{cudaMallocManaged()}.
However, with this method, the host memory space allocated is bound to the process which called the memory allocators.
Currently, due to the way how the CUDA memory allocators work, there is no way for the users to make the space allocated by them to be shareable with the other processes.
It is possible to create multiple copies of node feature tensors for each training process, but this leads to an extremely inefficient usage of host memory capacity.

\begin{lstlisting}[float,caption={Unified Tensor Declaration in Multiprocessing Environment},label={lst:multiprocessing}]
import torch
import torch.multiprocessing as mp

def producer(features, process_id, ...):
    <@\textcolor{blue}{\# Specify target GPU ID}@>
    torch.cuda.set_device(process_id)
    <@\textcolor{blue}{\# Map host shared memory to GPU address space}@>
    features = features.to(device="unified")
    ...
    
if __name__ == '__main__':
    features = torch.randn([100], device="cpu")
    <@\textcolor{blue}{\# Allocate shared memory space}@>
    features = features.share_memory_()
    ...
    <@\textcolor{blue}{\# Pass feature tensor alloctead in shared memory space}@>
    <@\textcolor{blue}{\# and call producers for multiple GPUs}@>    
    producer1 = ctx.Process(target=producer,
                            args=(features, 0, ...))
    producer2 = ctx.Process(target=producer,
                            args=(features, 1, ...))
    ...
\end{lstlisting}

Therefore, in our implementation, we take an opposite direction of memory allocation.
Instead of attempting to share a memory space after allocating with CUDA APIs, we first allocate a shareable memory space and then call CUDA APIs to allow GPUs access the space.
In Linux, to allow multiple processes to share a same memory space, \textit{shared memory} can be used.
Here, this \textit{shared memory} refers to a specific Linux implementation to allow interprocess communication and it should not be confused with other similar terminologies, such as the GPU shared memory.
We utilize the \texttt{cudaHostRegister()} API becasue it can be used on top of the Linux shared memory.
Therefore, by letting different processes to call \texttt{cudaHostRegister()} individually on the same shared memory space which has been already allocated, each GPU can get identical address mapping to the same host memory space.
The specific code that implements this approach is shown in Listing~\ref{lst:multiprocessing}. Line 14 shows the declaration of a shared memory tensor in the main process.
Shared memory allocation is already supported in PyTorch code by simply adding \texttt{.share\_memory\_()} command after the CPU tensor instance.
%{\color{red} WMH: But how about the unified tensor? Maybe say that the same command works for the unified tensor? Did we have to do anything special to enable this? It looks like we first allocate a CPU tensor and place it into the share memory. We then allow each GPU to convert the CPU tensor into a unified tensor. If this is true, we need to really clarify it. The code in Listing 3 does not quite make this design clear. Please check the accuracy of the sentences I added.}
%
To map this shared memory space for different GPUs, we pass the shared CPU tensor to the producer processes running on the GPUs (Lines 18 and 21). Each producer process simply calls the unified tensor declaration (Line 8) to effectively convert the shared CPU tensor into a unified tensor and maps it into the GPU's address space for zero-copy access.
Inside the producer code (Lines 4-9), the first thing that we must do is selecting the correct CUDA device (e.g. producer0 \textrightarrow ~ GPU0, producer1 \textrightarrow ~ GPU1, and so on).
Without this step, all unified tensor declarations in different producer processes will create a mapping for the default CUDA device defined by the system (e.g. GPU0).
\section{Evaluation}
\label{sec:eval}
This section presents an evaluation of the impact of our proposed design on GCN training time.
We first take a closer look of the improvements made by our optimizations one by one, and then show the overall training time reduction achieved.

\subsection{Methodology}
\subsubsection{Evaluation System}
For our evaluation, we use the system described in Table~\ref{tab:sysconfig}.
Our host system can hold two RTX 3090 GPUs and both are operating in PCIe 4.0 mode.
With PCIe 4.0 interconnects, both GPUs can achieve about 25.8GB/s of host to GPU DMA bandwidth in our microbenchmark.
The measured aggregated bandwidth of the two GPUs performing DMA on host memory at the same time is about 51.7GB/s.

\subsubsection{Application}
Our unified tensor implementation and the indexing kernel modification are based on PyTorch 1.8.0-nightly version.
For the GCN training, we use the GraphSAGE~\cite{hamilton2017inductive} implementation of DGL~\cite{wang2019dgl}.
We only modify the data communication portion of the implementation.
The sampling mechanism and the training algorithm remain unmodified.

\textbf{(a) CPU-Only} implementation only uses CPU for training GCN.
In this case, there is no need of data transfer over PCIe since GPUs are not involved in the training.

\textbf{(b) DMA-based} implementation uses CPU to gather node features into a contiguous buffer.
The gathered node features are transferred to GPUs by using DMA.

\textbf{(c) Na\"ive Zero-Copy} uses zero-copy as a main data transfer method, but do not include any optimizations we discussed in this paper.
Unified tensors are used to allow GPUs to perform zero-copy accesses on host memory.

\textbf{(c) Zero-Copy} implementation enables zero-copy accesses and additionally includes all optimizations we discussed in this paper.
Unified tensors are used to allow GPUs to perform zero-copy accesses on host memory.

\textbf{(d) All-in-GPU} implementation allocates the entire node feature array into each GPU memory before the training begins.
This implementation is used to show the rough upper bound of the performance improvement we can achieve through the data transfer optimization.
Due to the limited GPU memory capacity, we do not evaluate all datasets with this implementation.
We explicitly denote as "out-of-memory (OOM)" for such cases.

\subsubsection{Dataset}
In Table~\ref{tab:dataset}, we show the datasets we used for the evaluation.
\texttt{wikipedia}~\cite{konect} network consists of the wikilinks of the Wikipedia in the English language. Nodes are Wikipedia articles, and directed edges are wikilinks.
\texttt{amazon}~\cite{ni-etal-2019-justifying} dataset is based on Amazon product network connected by "also viewed" and "also bought" links.
\texttt{ogbn-papers100M} dataset is a directed citation graph of 111 million papers indexed by MAG~\cite{wang2020microsoft}.
The above datasets are used for basic performance evaluations.
\texttt{ogbn-products}~\cite{hu2020ogb} dataset is based on Amazon co-purchasing network~\cite{Bhatia16} where nodes represent products sold in Amazon, and edges between two products indicate that the products are purchased together.
\texttt{ogbn-products} is only used for the training time vs. node feature size sensitivity analysis on Section~\ref{sec:eval.nodesize}.

\begin{table}[t]
  \caption{Evaluation system configuration.}
  \label{tab:sysconfig}
  \begin{tabular}{cl}
    \toprule
        Category & \multicolumn{1}{c}{Specification} \\
    \midrule
        CPU & AMD Ryzen Threadripper 3960x 24C/48T\\
    \midrule
        Memory & DDR4 3200 MHz 256GB in Quad Channel\\
    \midrule
        GPU & 2x NVIDIA Ampere RTX 3090 24GB\\
    \midrule
        OS  & Ubuntu 20.04.1 \& Linux Kernel 5.8.0\\
    \midrule 
        S/W & CUDA 11.2 \& PyTorch 1.8.0-nightly\\
    \bottomrule
  \end{tabular}
\end{table}

\begin{table}[t]
  \caption{Evaluation Dataset.}
  \label{tab:dataset}
  \begin{tabular}{ccccc}
    \toprule
        Name & \#Feature & \#Node & \#Edge & Size \\
    \midrule
        ogbn-products & 128 - 4096 & 2.4M & 61.9M & -\\        
%    \midrule
        wikipedia & 315 & 13.6M & 437.2M & 17.1GB\\
%    \midrule
        amazon & 578 & 14.7M & 64.0M & 34.0GB\\
%    \midrule
        ogbn-papers100M & 128 & 111.1M & 1.6B & 56.9GB\\
    \bottomrule
  \end{tabular}
\end{table}

\begin{figure}[t]
  \centering
  \includegraphics[width=\linewidth]{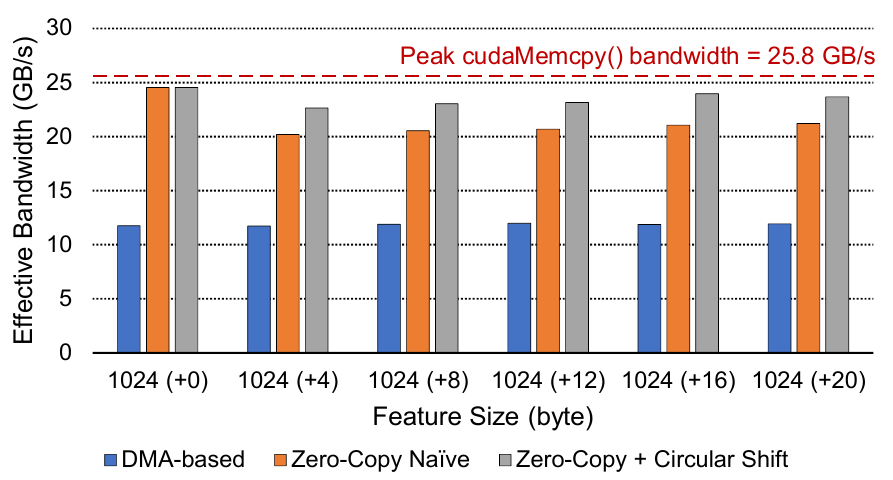}
  \caption{Effective data transfer bandwidth measured during the wikipedia dataset training. We sweep feature size to observe the impact of misaligned zero-copy accesses over PCIe.}
  \label{fig:alignment_measurement}
\end{figure}

\subsection{Bandwidth Analysis}
In Figure~\ref{fig:alignment_measurement}, we show the comparison of the effective bandwidths we measured during the wikipedia dataset training.
To observe the impact of the misaligned node feature access on the PCIe bandwidth, we sweep the node feature size from 1024B to 1044B in this experiment.
Zero-copy na\"ive approach does not implement the circular shift optimization we discussed in Section~\ref{sec:impl_aligned}.
Throughout the experiment, the effective bandwidth of the DMA-based approach is only about half of the zero-copy approaches as it requires a long CPU gathering process.

When the node feature size is 1024B, regardless of the circular shift optimization existence, the zero-copy implementations show the best effective bandwidth numbers.
Because the GPU cacheline size is 128B, in this case accessing any node features results in generating perfectly coalesced accesses.
Considering that the best \texttt{cudaMemcpy()} bandwidth we achieved is about 25.8GB/s, we can roughly estimate the upper bound efficiency of zero-copy access is about 95.1\%.
With more misaligned accesses, the efficiency of the na\"ive zero-copy implementation drops to 78-82\% while the optimized zero-copy implementation can achieve 88-93\% of efficiency.

In general, the results re-emphasize the importance of making cacheline-aligned accesses whenever using zero-copy accesses.
For savvy programmers, we expect them to understand the underlying hardware mechanism and to consider padding the input data if the overhead is not too big.
However, even if they fail to do so, our optimizations would still reduce the performance penalty for them.

\begin{figure}[t]
  \centering
  \includegraphics[width=\linewidth]{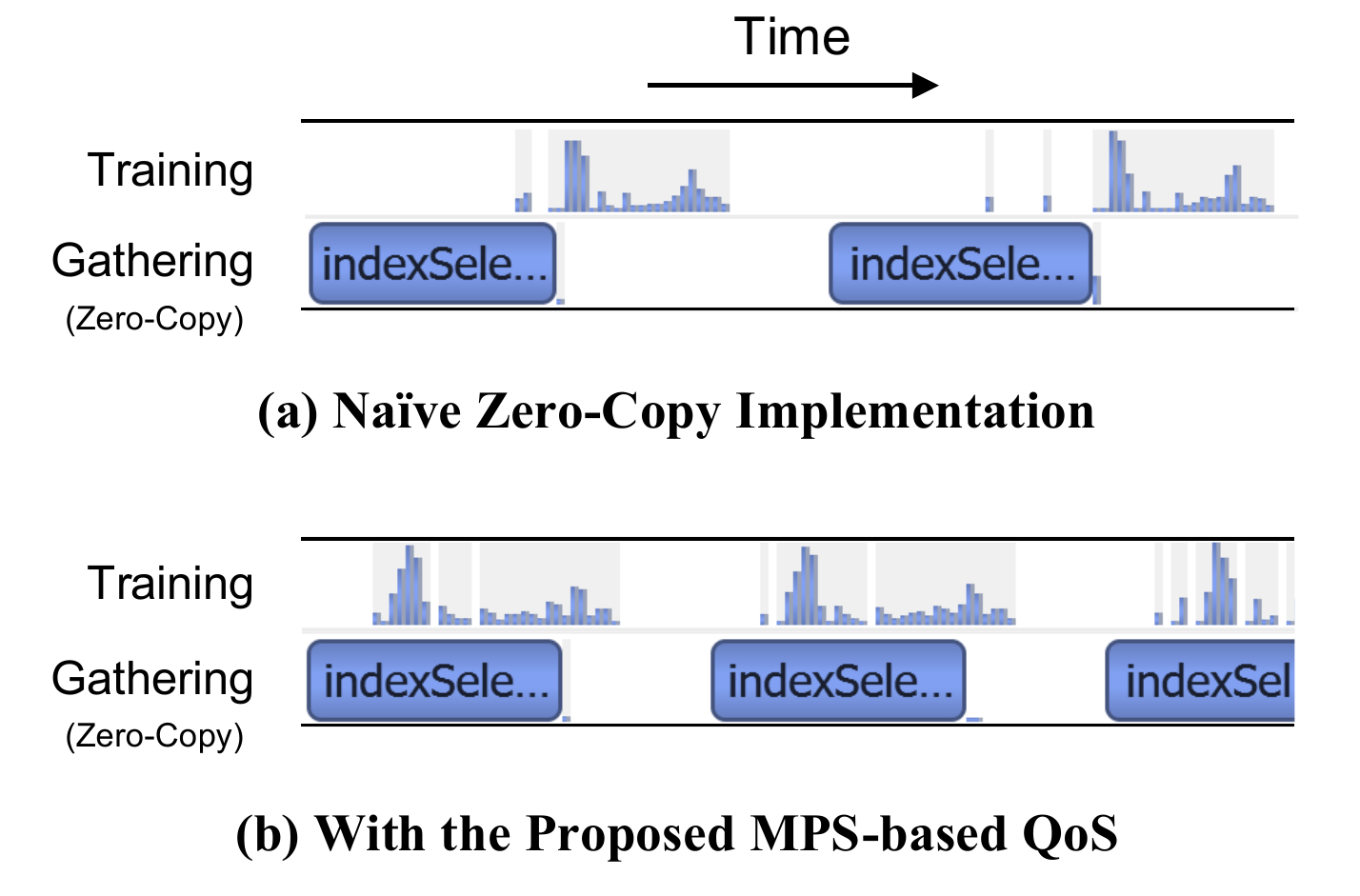}
  \caption{Snapshots of NVIDIA Nsight Systems Profiler during GCN training. CPU workloads not shown here.}
  \label{fig:mps_eval}
\end{figure}

\subsection{Concurrency Analysis}
The best way to check if our MPS-based resource provisioning is helping the concurrency is profiling the workload and visually inspecting the GPU kernel timeline.
In Figure~\ref{fig:mps_eval}, we show two profiling results of GCN training where (a) we do not apply any resource restriction and (b) we allocate 10\% of GPU resource for the zero-copy kernels and 90\% for the training kernels.
Without any MPS running, there is almost no concurrency occurring since each kernel is trying to consume the whole GPU resource.
In this specific case, the indexing (zero-copy) kernel is blocking other training kernels using GPU resources.
The training kernels are already scheduled into the queue, but most of them cannot be actually executed until the zero-copy kernel is finished.
Only a few kernels which require a small amount of GPU resource can be executed along the zero-copy kernel.
In the NVIDIA tools, the GPU is considered to be 100\% utilized at this point, but as we discussed in Section~\ref{sec:impl.async}, in fact only a limited number of cores can actually submit memory requests over PCIe due to the protocol limitation. Most of the cores are simply stalled, waiting for their turns to submit memory requests.
%
%Thus, most of the CUDA cores are idling in practice.
%

\begin{figure}[t]
  \centering
  \includegraphics[width=\linewidth]{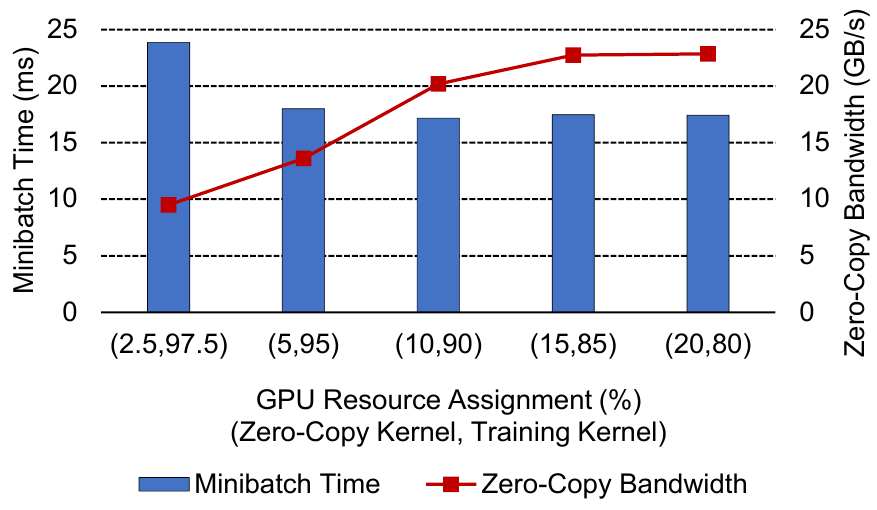}
  \caption{MPS resource partitioning ratio sensitivity analysis.}
  \label{fig:mps_bandwidth}
\end{figure}

On the other hand, when we enable the MPS and limit the the GPU resource usage for the zero-copy kernel to 10\%, it does not block the following training kernels anymore.
Furthermore, even though the zero-copy kernel can now use only up to 10\% of the GPU resource, there is no significant bandwidth drop caused from it.
In Figure~\ref{fig:mps_bandwidth}, we show the zero-copy PCIe bandwidth change over allocating different amount of GPU resource to the zero-copy kernel.
With 2.5--10\% of resource allocations, the zero-copy kernel cannot generate enough number of PCIe read requests and therefore the measured PCIe bandwidth is limited to 9.5--20.2GB/s.
Further increasing the GPU resource allocation can make the zero-copy bandwidth to reach around 23.0GB/s, but we do not observe any significant improvement after 15\% of allocation.
At this point, the amount of GPU resource allocated is excessive and we have already reached the maximum number of PCIe read requests that we can generate.
The results roughly fall in to the estimation we made in Section~\ref{sec:impl.async}.
If the other users want to apply the same optimization technique on different types of GPUs, the same methodology we used to make the estimation could be useful.

For the training time, 2.5--5\% of resource allocation is not enough to overlap (hide) the zero-copy kernel time with other processes and therefore the minibatch time is longer than the optimal case.
We achieve the best minibatch training time when the resource allocation is 10\%.
With more resource allocation on the zero-copy kernel, the computation kernels start to starve from lack of GPU resource.
If one wants to apply the same technique for different types of workloads, it might be worth to fine-tune the ratio.
However, still, one must be aware of the PCIe bandwidth limit.
For the rest of our evaluations, we simply use an allocation ratio of 10:90 since the minibatch time is quite stable with small variations in  the allocation ratio.
%, at least for our purpose, we do not see the need of fine-tuning.

\begin{figure}[t]
  \centering
  \includegraphics[width=\linewidth]{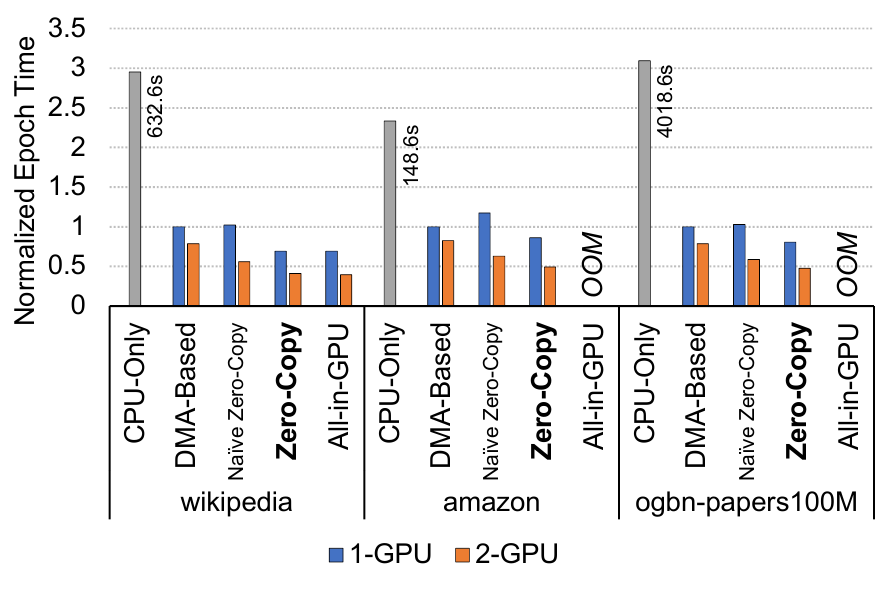}
  \caption{GCN training time comparison. OOM denotes out-of-memory.}
  \label{fig:eval_overall}
\end{figure}

\subsection{Training Performance}
\subsubsection{Overall Comparison}
In Figure~\ref{fig:eval_overall}, we show the overall training performance comparison.
Throughout the entire comparison, the CPU-only case shows the worst performance.
By limiting the computation unit to CPU, there is no need to worry about efficient data transfers over PCIe but at the same time the computation power is severely limited.
In general, the CPU-only method is 2.3--3.1$\times$ slower than the DMA-based method. This performance difference is an important motivation for moving the training part into the GPUs.

For the DMA-based method, doubling the number of GPUs does help reducing the overall training time, but an additional GPU results in only 21-27\% performance improvement across different datasets.
Because the DMA-based method make poor use of CPU resource and host memory bandwidth, increasing the number of workers quickly makes the entire training process to be throttled by them.
At this point, all GPUs are experiencing data starvation and have low utilizations.

For the na\"ive zero-copy method, we observe 2--17\% of performance degrade compared to the DMA-based method in a single-GPU setup.
This degradation is consistent with some of the conventional wisdoms that na\"ive zero-copy is inferior to DMA-based methods. Without our proposed optimizations, the zero-copy method suffers from the low bandwidth and the serialization issues described in Section~\ref{sec:impl_aligned} and Section~\ref{sec:impl.async}.
This result also gives us an idea how the programmers can make a premature conclusion to not further investigate the usage of zero-copy accesses.

With two GPUs, the na\"ive zero-copy method shows much better performance as well as performance scalability than the DMA-based method.
In a dual-GPU setup, the na\"ive zero-copy method becomes 30--41\% faster than the DMA-based method.
This is because, even without the optimizations, the zero-copy method by default much more efficiently uses the CPU resource and host memory bandwidth than the DMA-based method.
However, this benefit is not visible until the number of GPUs increases.

Finally, with our zero-copy optimizations, we can now clearly see the benefit of zero-copy in all cases.
In a single-GPU setup, the optimized zero-copy method is 16--44\% faster than the DMA-based method and in a dual-GPU setup, it is 65--92\% faster.
More surprisingly, with all optimizations included, the performance of the zero-copy method matches with the all-in-GPU method for the \texttt{wikipedia} dataset training.
Since the data communication time is completely hidden by the training process in this case, there is no disadvantage compared to the all-in-GPU method.
Overall, we observe a very significant benefit of using zero-copy accesses for GCN training.
Thanks to the design flow optimizations we discussed in Section~\ref{sec:impl.scheduling}, we do not observe any noticeable performance impact from the interprocess communications.

\subsubsection{Node Feature Size Sensitivity Analysis}
\label{sec:eval.nodesize}
Even though we use multiple different graphs to evaluate GCN training performance, some of other real-world datasets can have very different node feature dimensions.
For example, the node feature dimension of Pinterest dataset~\cite{ying2019pinsage} is about 4096, which is far larger than the node feature dimensions in our datasets.
However, many of those real-world datasets are proprietary and it is difficult to obtain for academic purposes.
Therefore, in this section, we artificially sweep the node feature dimension of \texttt{ogbn-products} dataset and compare the performances of zero-copy method and DMA-based method.

\begin{figure}[t]
  \centering
  \includegraphics[width=\linewidth]{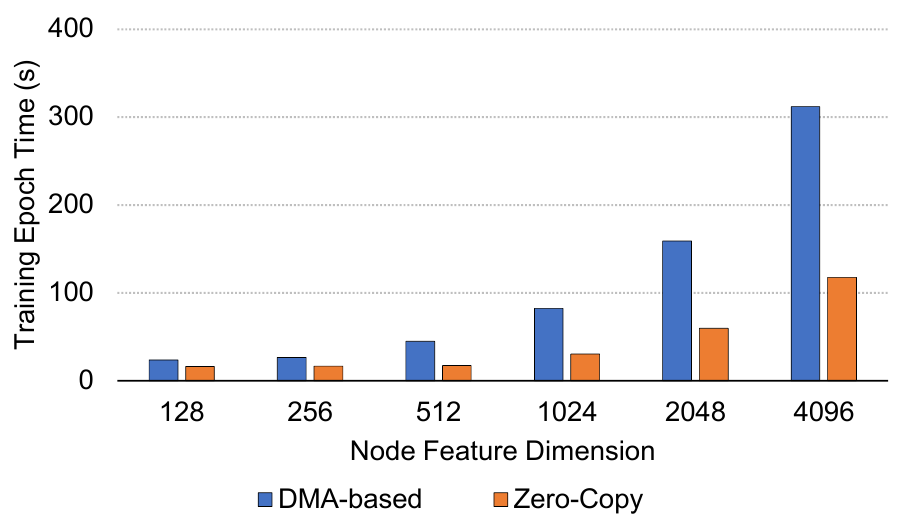}
  \caption{Node feature size sensitivity analysis.}
  \label{fig:eval_sweep}
\end{figure}

In Figure~\ref{fig:eval_sweep}, we show the node feature size sensitivity analysis results.
In this experiment, we use two GPUs.
With small node feature dimensions, the zero-copy method is only 1.5--1.6$\times$ faster than the DMA-based method.
However, when the node feature dimension is 4096, the zero-copy method is about 2.7$\times$ faster than the DMA-based method.
This is an expected behavior as with small node feature dimensions, other overheads in the GCN training processes take a sizeable portion of training time and therefore the data transfer time is relatively less important.
This experiment result stresses the necessity of using efficient data communication architecture when training GCN with large node feature dimensions.

\section{Discussion}
\label{sec:discuss}
\textbf{Unified Virtual Address (UVM) and Data Caching.} One benefit of using UVM is an automatic data caching with page migration.
As long as the migrated page remains in the GPU memory, it can quickly provide data accessed by GPU kernels without the need to go the host memory.
However, one must be aware of that even though the page migrations are assisted by NVIDIA hardware, still the significant portion of page managements and address translations are relying on slow software driver running on CPU.
In our work, due to the highly irregular nature of the graph-structure data, data caching is nearly useless.
By modifying our current zero-copy access implementation in PyTorch, we were able to additionally test UVM-based GCN training as well.
In average, we experienced about 40-50$\times$ of slowdown compared to the DMA-based method.
With the UVM method, both the CPU and GPU are completely throttled by migrating pages back and forth.
This suggests, that unless programmers are very familiar with a specific input dataset and they can apply certain dataset-specific locality optimizations, data caching is not an optimal solution.

\textbf{Physical GPU Resource Partitioning.}
Even though the CUDA MPS is already providing workload partitioning service, it is still a logical partitioning rather than a physical partitioning.
To guarantee a stable and high zero-copy PCIe bandwidth, it is better to physically isolate the GPU resources used by the training kernels from those used by the zero-copy kernels.
With recent introduction of Ampere architecture GPUs, NVIDIA started to support partitioning of a single GPU into multiple GPU instances~\cite{migguide}.
Each GPU instance has dedicated GPU memory resources which limit both the available capacity and bandwidth, and provide memory QoS.
Currently it is impossible to share a same piece of data between different GPU instances, but such advances in supporting hardware partitioning capability can potentially help isolating the zero-copy kernels with better QoS in the future.
\section{Related Works}
\label{sec:related}
\textbf{GCN Training on Very Large Graph.}
One of the most notable works of extreme-scale GCN training is PinSage~\cite{ying2019pinsage} which used a graph with 3 billions of nodes and 18 billions of edges.
In this work, multiple GPUs were used to accelerate the training process.
Similar to the DMA-based method that we described in our paper, PinSage also utilizes CPU to gather node features from the host side and then DMA to GPUs.
There are other works which desert GPUs and train only with CPUs due to the extreme memory capacity requirement.
SIGN~\cite{sign_icml_grl2020} uses GPUs in training, but the whole neighboring node aggregation steps are actually done by CPU.
%
%However, first of all, this work is more of an algorithm-oriented work rather than a performance-oriented work.
%
DistDGL~\cite{zheng2020distdgl} uses CPU-only distribute system to parallelize graph neural network training.
To manage the distributed storage, DistDGL requires complex data management processes to provide data in a timely fashion.
Results in our work show that having GPUs to perform zero-copy access to the host memory can offer major performance advantage over using only the CPUs.

\textbf{Alternative GCN Algorithms.} The out-of-GPU-memory issue can be also circumvented by modifying the application itself.
Several works~\cite{chiang2019cluster,graphsaint-iclr20,ma2019neugraph,jia2020improving} attempt to partition input graphs into smaller clusters prior to the training phase so each cluster can be fit into the GPU memory.
In this case, each training processor does not have view to the entire graph but only to the assigned clusters.
The immediate issue of this method is that partitioning graphs creates bias in the training result as it clusters similar nodes together~\cite{chiang2019cluster}.
Furthermore, partitioning graphs results in losing multiple edges which represent relational information~\cite{9046288}.
Especially with larger graphs, we need to create more number of partitions, and thus we also need to remove more number of edges during the partition.
It is empirically shown that the partitioning methods result in lower accuracy in several GCN training workloads~\cite{ogbn}.

\textbf{Other Graph-Related Workload Acceleration on GPU.} GCN training is not the only workload where the GPUs suffer from inefficient irregular host memory accesses.
There are several works which try to utilize GPUs in graph traversal workloads like PageRank~\cite{ilprints422} with large datasets.
Due to the usage of sparse matrix format for the representation of graph structure, traversing graphs results in generating very irregular memory accesses.
Considering that large graphs such as WDC14~\cite{wdc} has about 64 billions edges, the graphs cannot be placed in GPU memory and therefore the graph traversal workloads face the similar issue in this paper.
EMOGI~\cite{min2020emogi} utilizes zero-copy accesses to enable fine-grained host memory access during several graph traversal workloads.
Halo~\cite{Gera20} tries to ensure the spatial locality of graph nodes in the memory as well through extensive pre-processing.
However, the effectiveness of this method is completely random depending on the shape of the input graph.
Subway~\cite{Sabet20} uses a method very similar to the DMA-based method used in this work, which tries to utilize CPU as much as possible to gather scattered data for more efficient DMA.
\section{Conclusion}
\label{sec:conclusion}
In this work, we introduced a GPU-oriented, software defined data transfer architecture for efficient GCN training on large graphs.
In large-scale GCN training, one of the most challenging tasks is that how to efficiently transfer node features scattered in the host memory to GPUs.
As opposed to the traditional DMA-based method, we directly utilize GPU cores as a data moving agent to access sparse features in the host memory over zero-copy accesses.
Our evaluations show that together with zero-copy accesses and our optimizations, the GCN training performance can be improved by 65--92\%. Furthermore, the benefit of our proposed approach is significantly larger for 2-GPU training than 1-GPU training.
By implementing the end-to-end zero-copy based GCN training flow in PyTorch, we also show that our modifications can be seamlessly integrated with the existing high-level DNN programming models.

%\begin{acks}
%\end{acks}

%\clearpage

\balance
\bibliographystyle{ACM-Reference-Format}
\bibliography{sample}


\begin{thebibliography}{51}

%%% the Web address.

\ifx \showCODEN    \undefined \def \showCODEN     #1{\unskip}     \fi
\ifx \showDOI      \undefined \def \showDOI       #1{#1}\fi
\ifx \showISBNx    \undefined \def \showISBNx     #1{\unskip}     \fi
\ifx \showISBNxiii \undefined \def \showISBNxiii  #1{\unskip}     \fi
\ifx \showISSN     \undefined \def \showISSN      #1{\unskip}     \fi
\ifx \showLCCN     \undefined \def \showLCCN      #1{\unskip}     \fi
\ifx \shownote     \undefined \def \shownote      #1{#1}          \fi
\ifx \showarticletitle \undefined \def \showarticletitle #1{#1}   \fi
\ifx \showURL      \undefined \def \showURL       {\relax}        \fi
% invisible to TeX
\providecommand\bibfield[2]{#2}
\providecommand\bibinfo[2]{#2}
\providecommand\natexlab[1]{#1}
\providecommand\showeprint[2][]{arXiv:#2}

\bibitem[\protect\citeauthoryear{??}{ogb}{[n.d.]}]%
        {ogbn}
 \bibinfo{year}{[n.d.]}\natexlab{}.
\newblock \bibinfo{booktitle}{\emph{Leaderboards for Node Property
  Prediction}}.
\newblock
\urldef\tempurl%
\url{https://ogb.stanford.edu/docs/leader_nodeprop/}
\showURL{%
Retrieved Feb 21, 2021 from \tempurl}


\bibitem[\protect\citeauthoryear{??}{wdc}{[n.d.]}]%
        {wdc}
 \bibinfo{year}{[n.d.]}\natexlab{}.
\newblock \bibinfo{booktitle}{\emph{Web Data Commons - Hyperlink Graphs}}.
\newblock
\urldef\tempurl%
\url{http://webdatacommons.org/hyperlinkgraph/index.html}
\showURL{%
Retrieved Feb 21, 2021 from \tempurl}


\bibitem[\protect\citeauthoryear{Bhatia, Dahiya, Jain, Mittal, Prabhu, and
  Varma}{Bhatia et~al\mbox{.}}{2016}]%
        {Bhatia16}
\bibfield{author}{\bibinfo{person}{K. Bhatia}, \bibinfo{person}{K. Dahiya},
  \bibinfo{person}{H. Jain}, \bibinfo{person}{A. Mittal}, \bibinfo{person}{Y.
  Prabhu}, {and} \bibinfo{person}{M. Varma}.} \bibinfo{year}{2016}\natexlab{}.
\newblock \bibinfo{title}{The extreme classification repository: Multi-label
  datasets and code}.
\newblock
\newblock
\urldef\tempurl%
\url{http://manikvarma.org/downloads/XC/XMLRepository.html}
\showURL{%
\tempurl}


\bibitem[\protect\citeauthoryear{Bruna, Zaremba, Szlam, and Lecun}{Bruna
  et~al\mbox{.}}{2014}]%
        {lecunGCN}
\bibfield{author}{\bibinfo{person}{Joan Bruna}, \bibinfo{person}{Wojciech
  Zaremba}, \bibinfo{person}{Arthur Szlam}, {and} \bibinfo{person}{Yann
  Lecun}.} \bibinfo{year}{2014}\natexlab{}.
\newblock \showarticletitle{Spectral networks and locally connected networks on
  graphs}. In \bibinfo{booktitle}{\emph{International Conference on Learning
  Representations (ICLR2014), CBLS, April 2014}}.
\newblock


\bibitem[\protect\citeauthoryear{Chen, Ma, and Xiao}{Chen
  et~al\mbox{.}}{2018a}]%
        {chen2018fastgcn}
\bibfield{author}{\bibinfo{person}{Jie Chen}, \bibinfo{person}{Tengfei Ma},
  {and} \bibinfo{person}{Cao Xiao}.} \bibinfo{year}{2018}\natexlab{a}.
\newblock \showarticletitle{Fast{GCN}: Fast Learning with Graph Convolutional
  Networks via Importance Sampling}. In \bibinfo{booktitle}{\emph{International
  Conference on Learning Representations}}.
\newblock
\urldef\tempurl%
\url{https://openreview.net/forum?id=rytstxWAW}
\showURL{%
\tempurl}


\bibitem[\protect\citeauthoryear{Chen, Zhu, and Song}{Chen
  et~al\mbox{.}}{2018b}]%
        {pmlr-v80-chen18p}
\bibfield{author}{\bibinfo{person}{Jianfei Chen}, \bibinfo{person}{Jun Zhu},
  {and} \bibinfo{person}{Le Song}.} \bibinfo{year}{2018}\natexlab{b}.
\newblock \showarticletitle{Stochastic Training of Graph Convolutional Networks
  with Variance Reduction}. In \bibinfo{booktitle}{\emph{Proceedings of the
  35th International Conference on Machine Learning}}
  \emph{(\bibinfo{series}{Proceedings of Machine Learning Research})},
  \bibfield{editor}{\bibinfo{person}{Jennifer Dy} {and}
  \bibinfo{person}{Andreas Krause}} (Eds.), Vol.~\bibinfo{volume}{80}.
  \bibinfo{publisher}{PMLR}, \bibinfo{address}{Stockholmsmässan, Stockholm
  Sweden}, \bibinfo{pages}{942--950}.
\newblock
\urldef\tempurl%
\url{http://proceedings.mlr.press/v80/chen18p.html}
\showURL{%
\tempurl}


\bibitem[\protect\citeauthoryear{Chiang, Liu, Si, Li, Bengio, and Hsieh}{Chiang
  et~al\mbox{.}}{2019}]%
        {chiang2019cluster}
\bibfield{author}{\bibinfo{person}{Wei-Lin Chiang}, \bibinfo{person}{Xuanqing
  Liu}, \bibinfo{person}{Si Si}, \bibinfo{person}{Yang Li},
  \bibinfo{person}{Samy Bengio}, {and} \bibinfo{person}{Cho-Jui Hsieh}.}
  \bibinfo{year}{2019}\natexlab{}.
\newblock \showarticletitle{Cluster-GCN: An Efficient Algorithm for Training
  Deep and Large Graph Convolutional Networks}. In
  \bibinfo{booktitle}{\emph{Proceedings of the 25th ACM SIGKDD International
  Conference on Knowledge Discovery \& Data Mining}} (Anchorage, AK, USA)
  \emph{(\bibinfo{series}{KDD '19})}. \bibinfo{publisher}{Association for
  Computing Machinery}, \bibinfo{address}{New York, NY, USA},
  \bibinfo{pages}{257–266}.
\newblock
\showISBNx{9781450362016}
\urldef\tempurl%
\url{https://doi.org/10.1145/3292500.3330925}
\showDOI{\tempurl}


\bibitem[\protect\citeauthoryear{Defferrard, Bresson, and
  Vandergheynst}{Defferrard et~al\mbox{.}}{2016}]%
        {GCNPierre}
\bibfield{author}{\bibinfo{person}{Micha\"{e}l Defferrard},
  \bibinfo{person}{Xavier Bresson}, {and} \bibinfo{person}{Pierre
  Vandergheynst}.} \bibinfo{year}{2016}\natexlab{}.
\newblock \showarticletitle{Convolutional Neural Networks on Graphs with Fast
  Localized Spectral Filtering}. In \bibinfo{booktitle}{\emph{Proceedings of
  the 30th International Conference on Neural Information Processing Systems}}
  (Barcelona, Spain) \emph{(\bibinfo{series}{NIPS'16})}.
  \bibinfo{publisher}{Curran Associates Inc.}, \bibinfo{address}{Red Hook, NY,
  USA}, \bibinfo{pages}{3844–3852}.
\newblock
\showISBNx{9781510838819}


\bibitem[\protect\citeauthoryear{Fey and Lenssen}{Fey and Lenssen}{2019}]%
        {Fey/Lenssen/2019}
\bibfield{author}{\bibinfo{person}{Matthias Fey} {and} \bibinfo{person}{Jan~E.
  Lenssen}.} \bibinfo{year}{2019}\natexlab{}.
\newblock \showarticletitle{Fast Graph Representation Learning with {PyTorch
  Geometric}}. In \bibinfo{booktitle}{\emph{ICLR Workshop on Representation
  Learning on Graphs and Manifolds}}.
\newblock


\bibitem[\protect\citeauthoryear{Frasca, Rossi, Eynard, Chamberlain, Bronstein,
  and Monti}{Frasca et~al\mbox{.}}{2020}]%
        {sign_icml_grl2020}
\bibfield{author}{\bibinfo{person}{Fabrizio Frasca}, \bibinfo{person}{Emanuele
  Rossi}, \bibinfo{person}{Davide Eynard}, \bibinfo{person}{Benjamin
  Chamberlain}, \bibinfo{person}{Michael Bronstein}, {and}
  \bibinfo{person}{Federico Monti}.} \bibinfo{year}{2020}\natexlab{}.
\newblock \showarticletitle{SIGN: Scalable Inception Graph Neural Networks}. In
  \bibinfo{booktitle}{\emph{ICML 2020 Workshop on Graph Representation Learning
  and Beyond}}.
\newblock


\bibitem[\protect\citeauthoryear{Ganguly, Zhang, Yang, and Melhem}{Ganguly
  et~al\mbox{.}}{2020}]%
        {gangulyadaptive}
\bibfield{author}{\bibinfo{person}{Debashis Ganguly}, \bibinfo{person}{Z
  Zhang}, \bibinfo{person}{J Yang}, {and} \bibinfo{person}{Rami Melhem}.}
  \bibinfo{year}{2020}\natexlab{}.
\newblock \showarticletitle{Adaptive Page Migration for Irregular
  Data-intensive Applications under GPU Memory Oversubscription}. In
  \bibinfo{booktitle}{\emph{Proceedings of the Thirty-forth International
  Conference on Parallel and Distributed Processing (IPDPS)}}.
\newblock


\bibitem[\protect\citeauthoryear{Gera, Kim, Sao, Kim, and Bader}{Gera
  et~al\mbox{.}}{2020}]%
        {Gera20}
\bibfield{author}{\bibinfo{person}{Prasun Gera}, \bibinfo{person}{Hyojong Kim},
  \bibinfo{person}{Piyush Sao}, \bibinfo{person}{Hyesoon Kim}, {and}
  \bibinfo{person}{David Bader}.} \bibinfo{year}{2020}\natexlab{}.
\newblock \showarticletitle{Traversing Large Graphs on GPUs with Unified
  Memory}.
\newblock \bibinfo{journal}{\emph{Proceedings of the VLDB Endowment}}
  \bibinfo{volume}{13}, \bibinfo{number}{7} (\bibinfo{date}{March}
  \bibinfo{year}{2020}), \bibinfo{pages}{1119–1133}.
\newblock


\bibitem[\protect\citeauthoryear{Grover and Leskovec}{Grover and
  Leskovec}{2016}]%
        {node2vec}
\bibfield{author}{\bibinfo{person}{Aditya Grover} {and} \bibinfo{person}{Jure
  Leskovec}.} \bibinfo{year}{2016}\natexlab{}.
\newblock \showarticletitle{Node2vec: Scalable Feature Learning for Networks}.
  In \bibinfo{booktitle}{\emph{Proceedings of the 22nd ACM SIGKDD International
  Conference on Knowledge Discovery and Data Mining}} (San Francisco,
  California, USA) \emph{(\bibinfo{series}{KDD '16})}.
  \bibinfo{publisher}{Association for Computing Machinery},
  \bibinfo{address}{New York, NY, USA}, \bibinfo{pages}{855–864}.
\newblock
\showISBNx{9781450342322}
\urldef\tempurl%
\url{https://doi.org/10.1145/2939672.2939754}
\showDOI{\tempurl}


\bibitem[\protect\citeauthoryear{Hamilton, Ying, and Leskovec}{Hamilton
  et~al\mbox{.}}{2017a}]%
        {hamilton2017inductive}
\bibfield{author}{\bibinfo{person}{William~L. Hamilton}, \bibinfo{person}{Rex
  Ying}, {and} \bibinfo{person}{Jure Leskovec}.}
  \bibinfo{year}{2017}\natexlab{a}.
\newblock \showarticletitle{Inductive Representation Learning on Large Graphs}.
  In \bibinfo{booktitle}{\emph{Proceedings of the 31st International Conference
  on Neural Information Processing Systems}} (Long Beach, California, USA)
  \emph{(\bibinfo{series}{NIPS'17})}. \bibinfo{publisher}{Curran Associates
  Inc.}, \bibinfo{address}{Red Hook, NY, USA}, \bibinfo{pages}{1025–1035}.
\newblock
\showISBNx{9781510860964}


\bibitem[\protect\citeauthoryear{Hamilton, Ying, and Leskovec}{Hamilton
  et~al\mbox{.}}{2017b}]%
        {HamiltonYL17}
\bibfield{author}{\bibinfo{person}{William~L. Hamilton}, \bibinfo{person}{Rex
  Ying}, {and} \bibinfo{person}{Jure Leskovec}.}
  \bibinfo{year}{2017}\natexlab{b}.
\newblock \showarticletitle{Representation Learning on Graphs: Methods and
  Applications}.
\newblock \bibinfo{journal}{\emph{{IEEE} Data Eng. Bull.}}
  \bibinfo{volume}{40}, \bibinfo{number}{3} (\bibinfo{year}{2017}),
  \bibinfo{pages}{52--74}.
\newblock
\urldef\tempurl%
\url{http://sites.computer.org/debull/A17sept/p52.pdf}
\showURL{%
\tempurl}


\bibitem[\protect\citeauthoryear{Harris}{Harris}{2013}]%
        {howtocoalesce}
\bibfield{author}{\bibinfo{person}{Mark Harris}.}
  \bibinfo{year}{2013}\natexlab{}.
\newblock \bibinfo{booktitle}{\emph{How to Access Global Memory Efficiently in
  CUDA C/C++ Kernels}}.
\newblock
\urldef\tempurl%
\url{https://developer.nvidia.com/blog/how-access-global-memory-efficiently-cuda-c-kernels/}
\showURL{%
\tempurl}


\bibitem[\protect\citeauthoryear{Harris}{Harris}{2017}]%
        {UVMPrimer}
\bibfield{author}{\bibinfo{person}{Mark Harris}.}
  \bibinfo{year}{2017}\natexlab{}.
\newblock \bibinfo{booktitle}{\emph{Unified Memory for CUDA Beginners}}.
\newblock
\urldef\tempurl%
\url{https://developer.nvidia.com/blog/unified-memory-cuda-beginners/}
\showURL{%
\tempurl}


\bibitem[\protect\citeauthoryear{Hu, Fey, Zitnik, Dong, Ren, Liu, Catasta, and
  Leskovec}{Hu et~al\mbox{.}}{2020}]%
        {hu2020ogb}
\bibfield{author}{\bibinfo{person}{Weihua Hu}, \bibinfo{person}{Matthias Fey},
  \bibinfo{person}{Marinka Zitnik}, \bibinfo{person}{Yuxiao Dong},
  \bibinfo{person}{Hongyu Ren}, \bibinfo{person}{Bowen Liu},
  \bibinfo{person}{Michele Catasta}, {and} \bibinfo{person}{Jure Leskovec}.}
  \bibinfo{year}{2020}\natexlab{}.
\newblock \showarticletitle{Open Graph Benchmark: Datasets for Machine Learning
  on Graphs}.
\newblock \bibinfo{journal}{\emph{arXiv preprint arXiv:2005.00687}}
  (\bibinfo{year}{2020}).
\newblock


\bibitem[\protect\citeauthoryear{Jia, Lin, Gao, Zaharia, and Aiken}{Jia
  et~al\mbox{.}}{2020}]%
        {jia2020improving}
\bibfield{author}{\bibinfo{person}{Zhihao Jia}, \bibinfo{person}{Sina Lin},
  \bibinfo{person}{Mingyu Gao}, \bibinfo{person}{Matei Zaharia}, {and}
  \bibinfo{person}{Alex Aiken}.} \bibinfo{year}{2020}\natexlab{}.
\newblock \showarticletitle{Improving the accuracy, scalability, and
  performance of graph neural networks with Roc}.
\newblock \bibinfo{journal}{\emph{Proceedings of Machine Learning and Systems
  (MLSys)}} (\bibinfo{year}{2020}), \bibinfo{pages}{187--198}.
\newblock


\bibitem[\protect\citeauthoryear{Kipf and Welling}{Kipf and Welling}{2016}]%
        {kipf2016variational}
\bibfield{author}{\bibinfo{person}{Thomas~N Kipf} {and} \bibinfo{person}{Max
  Welling}.} \bibinfo{year}{2016}\natexlab{}.
\newblock \showarticletitle{Variational Graph Auto-Encoders}.
\newblock \bibinfo{journal}{\emph{NIPS Workshop on Bayesian Deep Learning}}
  (\bibinfo{year}{2016}).
\newblock


\bibitem[\protect\citeauthoryear{Kipf and Welling}{Kipf and Welling}{2017}]%
        {kipf2017semi}
\bibfield{author}{\bibinfo{person}{Thomas~N. Kipf} {and} \bibinfo{person}{Max
  Welling}.} \bibinfo{year}{2017}\natexlab{}.
\newblock \showarticletitle{Semi-Supervised Classification with Graph
  Convolutional Networks}. In \bibinfo{booktitle}{\emph{International
  Conference on Learning Representations (ICLR)}}.
\newblock


\bibitem[\protect\citeauthoryear{Kunegis}{Kunegis}{2013}]%
        {konect}
\bibfield{author}{\bibinfo{person}{J\'{e}r\^{o}me Kunegis}.}
  \bibinfo{year}{2013}\natexlab{}.
\newblock \showarticletitle{KONECT: The Koblenz Network Collection}. In
  \bibinfo{booktitle}{\emph{Proceedings of the 22nd International Conference on
  World Wide Web}} (Rio de Janeiro, Brazil) \emph{(\bibinfo{series}{WWW '13
  Companion})}. \bibinfo{publisher}{Association for Computing Machinery},
  \bibinfo{address}{New York, NY, USA}, \bibinfo{pages}{1343–1350}.
\newblock
\showISBNx{9781450320382}
\urldef\tempurl%
\url{https://doi.org/10.1145/2487788.2488173}
\showDOI{\tempurl}


\bibitem[\protect\citeauthoryear{{LeCun}, {Boser}, {Denker}, {Henderson},
  {Howard}, {Hubbard}, and {Jackel}}{{LeCun} et~al\mbox{.}}{1989}]%
        {CNN0}
\bibfield{author}{\bibinfo{person}{Y. {LeCun}}, \bibinfo{person}{B. {Boser}},
  \bibinfo{person}{J.~S. {Denker}}, \bibinfo{person}{D. {Henderson}},
  \bibinfo{person}{R.~E. {Howard}}, \bibinfo{person}{W. {Hubbard}}, {and}
  \bibinfo{person}{L.~D. {Jackel}}.} \bibinfo{year}{1989}\natexlab{}.
\newblock \showarticletitle{Backpropagation Applied to Handwritten Zip Code
  Recognition}.
\newblock \bibinfo{journal}{\emph{Neural Computation}} \bibinfo{volume}{1},
  \bibinfo{number}{4} (\bibinfo{year}{1989}), \bibinfo{pages}{541--551}.
\newblock
\urldef\tempurl%
\url{https://doi.org/10.1162/neco.1989.1.4.541}
\showDOI{\tempurl}


\bibitem[\protect\citeauthoryear{Lin, Han, Mao, Wang, and Dally}{Lin
  et~al\mbox{.}}{2018}]%
        {lin2018deep}
\bibfield{author}{\bibinfo{person}{Yujun Lin}, \bibinfo{person}{Song Han},
  \bibinfo{person}{Huizi Mao}, \bibinfo{person}{Yu Wang}, {and}
  \bibinfo{person}{Bill Dally}.} \bibinfo{year}{2018}\natexlab{}.
\newblock \showarticletitle{Deep Gradient Compression: Reducing the
  Communication Bandwidth for Distributed Training}. In
  \bibinfo{booktitle}{\emph{International Conference on Learning
  Representations}}.
\newblock
\urldef\tempurl%
\url{https://openreview.net/forum?id=SkhQHMW0W}
\showURL{%
\tempurl}


\bibitem[\protect\citeauthoryear{Ma, Yang, Miao, Xue, Wu, Zhou, and Dai}{Ma
  et~al\mbox{.}}{2019}]%
        {ma2019neugraph}
\bibfield{author}{\bibinfo{person}{Lingxiao Ma}, \bibinfo{person}{Zhi Yang},
  \bibinfo{person}{Youshan Miao}, \bibinfo{person}{Jilong Xue},
  \bibinfo{person}{Ming Wu}, \bibinfo{person}{Lidong Zhou}, {and}
  \bibinfo{person}{Yafei Dai}.} \bibinfo{year}{2019}\natexlab{}.
\newblock \showarticletitle{Neugraph: Parallel Deep Neural Network Computation
  on Large Graphs}. In \bibinfo{booktitle}{\emph{Proceedings of the 2019 USENIX
  Conference on Usenix Annual Technical Conference}} (Renton, WA, USA)
  \emph{(\bibinfo{series}{USENIX ATC '19})}. \bibinfo{publisher}{USENIX
  Association}, \bibinfo{address}{USA}, \bibinfo{pages}{443–457}.
\newblock
\showISBNx{9781939133038}


\bibitem[\protect\citeauthoryear{Min, Mailthody, Qureshi, Xiong, Ebrahimi, and
  Hwu}{Min et~al\mbox{.}}{2020}]%
        {min2020emogi}
\bibfield{author}{\bibinfo{person}{Seung~Won Min},
  \bibinfo{person}{Vikram~Sharma Mailthody}, \bibinfo{person}{Zaid Qureshi},
  \bibinfo{person}{Jinjun Xiong}, \bibinfo{person}{Eiman Ebrahimi}, {and}
  \bibinfo{person}{Wen-mei Hwu}.} \bibinfo{year}{2020}\natexlab{}.
\newblock \showarticletitle{EMOGI: Efficient Memory-access for Out-of-memory
  Graph-traversal In GPUs}.
\newblock \bibinfo{journal}{\emph{arXiv preprint arXiv:2006.06890}}
  (\bibinfo{year}{2020}).
\newblock


\bibitem[\protect\citeauthoryear{Neugebauer, Antichi, Zazo, Audzevich,
  L\'{o}pez-Buedo, and Moore}{Neugebauer et~al\mbox{.}}{2018}]%
        {10.1145/3230543.3230560}
\bibfield{author}{\bibinfo{person}{Rolf Neugebauer}, \bibinfo{person}{Gianni
  Antichi}, \bibinfo{person}{Jos\'{e}~Fernando Zazo}, \bibinfo{person}{Yury
  Audzevich}, \bibinfo{person}{Sergio L\'{o}pez-Buedo}, {and}
  \bibinfo{person}{Andrew~W. Moore}.} \bibinfo{year}{2018}\natexlab{}.
\newblock \showarticletitle{Understanding PCIe Performance for End Host
  Networking}. In \bibinfo{booktitle}{\emph{Proceedings of the 2018 Conference
  of the ACM Special Interest Group on Data Communication}} (Budapest, Hungary)
  \emph{(\bibinfo{series}{SIGCOMM '18})}. \bibinfo{publisher}{Association for
  Computing Machinery}, \bibinfo{address}{New York, NY, USA},
  \bibinfo{pages}{327–341}.
\newblock
\showISBNx{9781450355674}
\urldef\tempurl%
\url{https://doi.org/10.1145/3230543.3230560}
\showDOI{\tempurl}


\bibitem[\protect\citeauthoryear{Ni, Li, and McAuley}{Ni et~al\mbox{.}}{2019}]%
        {ni-etal-2019-justifying}
\bibfield{author}{\bibinfo{person}{Jianmo Ni}, \bibinfo{person}{Jiacheng Li},
  {and} \bibinfo{person}{Julian McAuley}.} \bibinfo{year}{2019}\natexlab{}.
\newblock \showarticletitle{Justifying Recommendations using Distantly-Labeled
  Reviews and Fine-Grained Aspects}. In \bibinfo{booktitle}{\emph{Proceedings
  of the 2019 Conference on Empirical Methods in Natural Language Processing
  and the 9th International Joint Conference on Natural Language Processing
  (EMNLP-IJCNLP)}}. \bibinfo{publisher}{Association for Computational
  Linguistics}, \bibinfo{address}{Hong Kong, China}, \bibinfo{pages}{188--197}.
\newblock
\urldef\tempurl%
\url{https://doi.org/10.18653/v1/D19-1018}
\showDOI{\tempurl}


\bibitem[\protect\citeauthoryear{Niepert, Ahmed, and Kutzkov}{Niepert
  et~al\mbox{.}}{2016}]%
        {pmlr-v48-niepert16}
\bibfield{author}{\bibinfo{person}{Mathias Niepert}, \bibinfo{person}{Mohamed
  Ahmed}, {and} \bibinfo{person}{Konstantin Kutzkov}.}
  \bibinfo{year}{2016}\natexlab{}.
\newblock \showarticletitle{Learning Convolutional Neural Networks for Graphs}.
  In \bibinfo{booktitle}{\emph{Proceedings of The 33rd International Conference
  on Machine Learning}} \emph{(\bibinfo{series}{Proceedings of Machine Learning
  Research})}, \bibfield{editor}{\bibinfo{person}{Maria~Florina Balcan} {and}
  \bibinfo{person}{Kilian~Q. Weinberger}} (Eds.), Vol.~\bibinfo{volume}{48}.
  \bibinfo{publisher}{PMLR}, \bibinfo{address}{New York, New York, USA},
  \bibinfo{pages}{2014--2023}.
\newblock
\urldef\tempurl%
\url{http://proceedings.mlr.press/v48/niepert16.html}
\showURL{%
\tempurl}


\bibitem[\protect\citeauthoryear{Nvidia}{Nvidia}{2016}]%
        {P100Whitepaper}
\bibfield{author}{\bibinfo{person}{Nvidia}.} \bibinfo{year}{2016}\natexlab{}.
\newblock \bibinfo{booktitle}{\emph{Nvidia Tesla P100 Whitepaper}}.
\newblock
\urldef\tempurl%
\url{https://images.nvidia.com/content/pdf/tesla/whitepaper/pascal-architecture-whitepaper.pdf}
\showURL{%
\tempurl}


\bibitem[\protect\citeauthoryear{Nvidia}{Nvidia}{2017}]%
        {V100Whitepaper}
\bibfield{author}{\bibinfo{person}{Nvidia}.} \bibinfo{year}{2017}\natexlab{}.
\newblock \bibinfo{booktitle}{\emph{Nvidia Tesla V100 GPU Architecture
  Whitepaper}}.
\newblock
\urldef\tempurl%
\url{https://images.nvidia.com/content/volta-architecture/pdf/volta-architecture-whitepaper.pdf}
\showURL{%
\tempurl}


\bibitem[\protect\citeauthoryear{NVIDIA}{NVIDIA}{2020}]%
        {mpsguide}
\bibfield{author}{\bibinfo{person}{NVIDIA}.} \bibinfo{year}{2020}\natexlab{}.
\newblock \bibinfo{title}{{MULTI-PROCESS SERVICE}}.
\newblock
\newblock
\urldef\tempurl%
\url{https://docs.nvidia.com/deploy/pdf/CUDA_Multi_Process_Service_Overview.pdf}
\showURL{%
\tempurl}


\bibitem[\protect\citeauthoryear{Nvidia}{Nvidia}{2020}]%
        {A100Whitepaper}
\bibfield{author}{\bibinfo{person}{Nvidia}.} \bibinfo{year}{2020}\natexlab{}.
\newblock \bibinfo{booktitle}{\emph{Nvidia A100 TensorCore GPU Architecture
  Whitepaper}}.
\newblock
\urldef\tempurl%
\url{https://www.nvidia.com/content/dam/en-zz/Solutions/Data-Center/nvidia-ampere-architecture-whitepaper.pdf}
\showURL{%
\tempurl}


\bibitem[\protect\citeauthoryear{NVIDIA}{NVIDIA}{2020}]%
        {migguide}
\bibfield{author}{\bibinfo{person}{NVIDIA}.} \bibinfo{year}{2020}\natexlab{}.
\newblock \bibinfo{title}{{NVIDIA MULTI-INSTANCE GPU USERGUIDE}}.
\newblock
\newblock
\urldef\tempurl%
\url{https://docs.nvidia.com/datacenter/tesla/pdf/NVIDIA_MIG_User_Guide.pdf}
\showURL{%
\tempurl}


\bibitem[\protect\citeauthoryear{NVIDIA}{NVIDIA}{2021}]%
        {profileguide}
\bibfield{author}{\bibinfo{person}{NVIDIA}.} \bibinfo{year}{2021}\natexlab{}.
\newblock \bibinfo{title}{{KERNEL PROFILING GUIDE}}.
\newblock
\newblock
\urldef\tempurl%
\url{https://docs.nvidia.com/nsight-compute/pdf/ProfilingGuide.pdf}
\showURL{%
\tempurl}


\bibitem[\protect\citeauthoryear{Page, Brin, Motwani, and Winograd}{Page
  et~al\mbox{.}}{1999}]%
        {ilprints422}
\bibfield{author}{\bibinfo{person}{Lawrence Page}, \bibinfo{person}{Sergey
  Brin}, \bibinfo{person}{Rajeev Motwani}, {and} \bibinfo{person}{Terry
  Winograd}.} \bibinfo{year}{1999}\natexlab{}.
\newblock \bibinfo{booktitle}{\emph{The PageRank Citation Ranking: Bringing
  Order to the Web.}}
\newblock \bibinfo{type}{Technical Report} 1999-66.
  \bibinfo{institution}{Stanford InfoLab}.
\newblock
\urldef\tempurl%
\url{http://ilpubs.stanford.edu:8090/422/}
\showURL{%
\tempurl}
\newblock
\shownote{Previous number = SIDL-WP-1999-0120.}


\bibitem[\protect\citeauthoryear{Pearson, Dakkak, Hashash, Li, Chung, Xiong,
  and Hwu}{Pearson et~al\mbox{.}}{2019}]%
        {pearson19}
\bibfield{author}{\bibinfo{person}{Carl Pearson}, \bibinfo{person}{Abdul
  Dakkak}, \bibinfo{person}{Sarah Hashash}, \bibinfo{person}{Cheng Li},
  \bibinfo{person}{I-Hsin Chung}, \bibinfo{person}{Jinjun Xiong}, {and}
  \bibinfo{person}{Wen-Mei Hwu}.} \bibinfo{year}{2019}\natexlab{}.
\newblock \showarticletitle{Evaluating Characteristics of CUDA Communication
  Primitives on High-Bandwidth Interconnects}. In
  \bibinfo{booktitle}{\emph{Proceedings of the 2019 ACM/SPEC International
  Conference on Performance Engineering}} (Mumbai, India)
  \emph{(\bibinfo{series}{ICPE '19})}. \bibinfo{publisher}{Association for
  Computing Machinery}, \bibinfo{address}{New York, NY, USA},
  \bibinfo{pages}{209–218}.
\newblock
\showISBNx{9781450362399}
\urldef\tempurl%
\url{https://doi.org/10.1145/3297663.3310299}
\showDOI{\tempurl}


\bibitem[\protect\citeauthoryear{Perozzi, Al-Rfou, and Skiena}{Perozzi
  et~al\mbox{.}}{2014}]%
        {DeepWalk}
\bibfield{author}{\bibinfo{person}{Bryan Perozzi}, \bibinfo{person}{Rami
  Al-Rfou}, {and} \bibinfo{person}{Steven Skiena}.}
  \bibinfo{year}{2014}\natexlab{}.
\newblock \showarticletitle{DeepWalk: Online Learning of Social
  Representations}. In \bibinfo{booktitle}{\emph{Proceedings of the 20th ACM
  SIGKDD International Conference on Knowledge Discovery and Data Mining}} (New
  York, New York, USA) \emph{(\bibinfo{series}{KDD '14})}.
  \bibinfo{publisher}{Association for Computing Machinery},
  \bibinfo{address}{New York, NY, USA}, \bibinfo{pages}{701–710}.
\newblock
\showISBNx{9781450329569}
\urldef\tempurl%
\url{https://doi.org/10.1145/2623330.2623732}
\showDOI{\tempurl}


\bibitem[\protect\citeauthoryear{{Rhu}, {Gimelshein}, {Clemons}, {Zulfiqar},
  and {Keckler}}{{Rhu} et~al\mbox{.}}{2016}]%
        {7783721}
\bibfield{author}{\bibinfo{person}{M. {Rhu}}, \bibinfo{person}{N.
  {Gimelshein}}, \bibinfo{person}{J. {Clemons}}, \bibinfo{person}{A.
  {Zulfiqar}}, {and} \bibinfo{person}{S.~W. {Keckler}}.}
  \bibinfo{year}{2016}\natexlab{}.
\newblock \showarticletitle{vDNN: Virtualized deep neural networks for
  scalable, memory-efficient neural network design}. In
  \bibinfo{booktitle}{\emph{2016 49th Annual IEEE/ACM International Symposium
  on Microarchitecture (MICRO)}}. \bibinfo{pages}{1--13}.
\newblock
\urldef\tempurl%
\url{https://doi.org/10.1109/MICRO.2016.7783721}
\showDOI{\tempurl}


\bibitem[\protect\citeauthoryear{Sabet, Zhao, and Gupta}{Sabet
  et~al\mbox{.}}{2020}]%
        {Sabet20}
\bibfield{author}{\bibinfo{person}{Amir Hossein~Nodehi Sabet},
  \bibinfo{person}{Zhijia Zhao}, {and} \bibinfo{person}{Rajiv Gupta}.}
  \bibinfo{year}{2020}\natexlab{}.
\newblock \showarticletitle{Subway: Minimizing Data Transfer during
  out-of-GPU-Memory Graph Processing}. In \bibinfo{booktitle}{\emph{Proceedings
  of the Fifteenth European Conference on Computer Systems}} (Heraklion,
  Greece) \emph{(\bibinfo{series}{EuroSys ’20})}.
  \bibinfo{publisher}{Association for Computing Machinery},
  \bibinfo{address}{New York, NY, USA}, Article \bibinfo{articleno}{12},
  \bibinfo{numpages}{16}~pages.
\newblock


\bibitem[\protect\citeauthoryear{Schroeder}{Schroeder}{2011}]%
        {NvidiaUVA}
\bibfield{author}{\bibinfo{person}{Tim Schroeder}.}
  \bibinfo{year}{2011}\natexlab{}.
\newblock \bibinfo{booktitle}{\emph{Peer-to-Peer \& Unified Virtual
  Addressing}}.
\newblock
\urldef\tempurl%
\url{https://developer.download.nvidia.com/CUDA/training/cuda_webinars_GPUDirect_uva.pdf}
\showURL{%
\tempurl}


\bibitem[\protect\citeauthoryear{Strom}{Strom}{2015}]%
        {strom2015scalable}
\bibfield{author}{\bibinfo{person}{Nikko Strom}.}
  \bibinfo{year}{2015}\natexlab{}.
\newblock \showarticletitle{Scalable distributed DNN training using commodity
  GPU cloud computing}. In \bibinfo{booktitle}{\emph{Sixteenth Annual
  Conference of the International Speech Communication Association}}.
\newblock


\bibitem[\protect\citeauthoryear{{Sze}, {Chen}, {Yang}, and {Emer}}{{Sze}
  et~al\mbox{.}}{2017}]%
        {8114708}
\bibfield{author}{\bibinfo{person}{V. {Sze}}, \bibinfo{person}{Y. {Chen}},
  \bibinfo{person}{T. {Yang}}, {and} \bibinfo{person}{J.~S. {Emer}}.}
  \bibinfo{year}{2017}\natexlab{}.
\newblock \showarticletitle{Efficient Processing of Deep Neural Networks: A
  Tutorial and Survey}.
\newblock \bibinfo{journal}{\emph{Proc. IEEE}} \bibinfo{volume}{105},
  \bibinfo{number}{12} (\bibinfo{year}{2017}), \bibinfo{pages}{2295--2329}.
\newblock
\urldef\tempurl%
\url{https://doi.org/10.1109/JPROC.2017.2761740}
\showDOI{\tempurl}


\bibitem[\protect\citeauthoryear{Volkov}{Volkov}{2016}]%
        {volkov2016understanding}
\bibfield{author}{\bibinfo{person}{Vasily Volkov}.}
  \bibinfo{year}{2016}\natexlab{}.
\newblock \emph{\bibinfo{title}{Understanding latency hiding on GPUs}}.
\newblock \bibinfo{thesistype}{Ph.D. Dissertation}. \bibinfo{school}{UC
  Berkeley}.
\newblock


\bibitem[\protect\citeauthoryear{Wang, Shen, Huang, Wu, Dong, and Kanakia}{Wang
  et~al\mbox{.}}{2020}]%
        {wang2020microsoft}
\bibfield{author}{\bibinfo{person}{Kuansan Wang}, \bibinfo{person}{Iris Shen},
  \bibinfo{person}{Charles Huang}, \bibinfo{person}{Chieh-Han Wu},
  \bibinfo{person}{Yuxiao Dong}, {and} \bibinfo{person}{Anshul Kanakia}.}
  \bibinfo{year}{2020}\natexlab{}.
\newblock \showarticletitle{Microsoft Academic Graph: when experts are not
  enough}.
\newblock \bibinfo{journal}{\emph{Quantitative Science Studies}}
  \bibinfo{volume}{1}, \bibinfo{number}{1} (\bibinfo{date}{February}
  \bibinfo{year}{2020}), \bibinfo{pages}{396--413}.
\newblock
\newblock
\shownote{https://doi.org/10.1162/qss\_a\_00021.}


\bibitem[\protect\citeauthoryear{Wang, Ye, Zhao, Wu, Li, Song, Xu, and
  Kraska}{Wang et~al\mbox{.}}{2018}]%
        {10.1145/3200691.3178491}
\bibfield{author}{\bibinfo{person}{Linnan Wang}, \bibinfo{person}{Jinmian Ye},
  \bibinfo{person}{Yiyang Zhao}, \bibinfo{person}{Wei Wu}, \bibinfo{person}{Ang
  Li}, \bibinfo{person}{Shuaiwen~Leon Song}, \bibinfo{person}{Zenglin Xu},
  {and} \bibinfo{person}{Tim Kraska}.} \bibinfo{year}{2018}\natexlab{}.
\newblock \showarticletitle{Superneurons: Dynamic GPU Memory Management for
  Training Deep Neural Networks}.
\newblock \bibinfo{journal}{\emph{SIGPLAN Not.}} \bibinfo{volume}{53},
  \bibinfo{number}{1} (\bibinfo{date}{Feb.} \bibinfo{year}{2018}),
  \bibinfo{pages}{41–53}.
\newblock
\showISSN{0362-1340}
\urldef\tempurl%
\url{https://doi.org/10.1145/3200691.3178491}
\showDOI{\tempurl}


\bibitem[\protect\citeauthoryear{Wang, Zheng, Ye, Gan, Li, Song, Zhou, Ma, Yu,
  Gai, Xiao, He, Karypis, Li, and Zhang}{Wang et~al\mbox{.}}{2019}]%
        {wang2019dgl}
\bibfield{author}{\bibinfo{person}{Minjie Wang}, \bibinfo{person}{Da Zheng},
  \bibinfo{person}{Zihao Ye}, \bibinfo{person}{Quan Gan},
  \bibinfo{person}{Mufei Li}, \bibinfo{person}{Xiang Song},
  \bibinfo{person}{Jinjing Zhou}, \bibinfo{person}{Chao Ma},
  \bibinfo{person}{Lingfan Yu}, \bibinfo{person}{Yu Gai},
  \bibinfo{person}{Tianjun Xiao}, \bibinfo{person}{Tong He},
  \bibinfo{person}{George Karypis}, \bibinfo{person}{Jinyang Li}, {and}
  \bibinfo{person}{Zheng Zhang}.} \bibinfo{year}{2019}\natexlab{}.
\newblock \showarticletitle{Deep Graph Library: A Graph-Centric,
  Highly-Performant Package for Graph Neural Networks}.
\newblock \bibinfo{journal}{\emph{arXiv preprint arXiv:1909.01315}}
  (\bibinfo{year}{2019}).
\newblock


\bibitem[\protect\citeauthoryear{Wu, Pan, Chen, Long, Zhang, and Yu}{Wu
  et~al\mbox{.}}{2021}]%
        {9046288}
\bibfield{author}{\bibinfo{person}{Zonghan Wu}, \bibinfo{person}{Shirui Pan},
  \bibinfo{person}{Fengwen Chen}, \bibinfo{person}{Guodong Long},
  \bibinfo{person}{Chengqi Zhang}, {and} \bibinfo{person}{Philip~S. Yu}.}
  \bibinfo{year}{2021}\natexlab{}.
\newblock \showarticletitle{A Comprehensive Survey on Graph Neural Networks}.
\newblock \bibinfo{journal}{\emph{IEEE Transactions on Neural Networks and
  Learning Systems}} \bibinfo{volume}{32}, \bibinfo{number}{1}
  (\bibinfo{year}{2021}), \bibinfo{pages}{4--24}.
\newblock
\urldef\tempurl%
\url{https://doi.org/10.1109/TNNLS.2020.2978386}
\showDOI{\tempurl}


\bibitem[\protect\citeauthoryear{Ying, He, Chen, Eksombatchai, Hamilton, and
  Leskovec}{Ying et~al\mbox{.}}{2018}]%
        {ying2019pinsage}
\bibfield{author}{\bibinfo{person}{Rex Ying}, \bibinfo{person}{Ruining He},
  \bibinfo{person}{Kaifeng Chen}, \bibinfo{person}{Pong Eksombatchai},
  \bibinfo{person}{William~L. Hamilton}, {and} \bibinfo{person}{Jure
  Leskovec}.} \bibinfo{year}{2018}\natexlab{}.
\newblock \showarticletitle{Graph Convolutional Neural Networks for Web-Scale
  Recommender Systems}. In \bibinfo{booktitle}{\emph{Proceedings of the 24th
  ACM SIGKDD International Conference on Knowledge Discovery \& Data Mining}}
  (London, United Kingdom) \emph{(\bibinfo{series}{KDD '18})}.
  \bibinfo{publisher}{Association for Computing Machinery},
  \bibinfo{address}{New York, NY, USA}, \bibinfo{pages}{974–983}.
\newblock
\showISBNx{9781450355520}
\urldef\tempurl%
\url{https://doi.org/10.1145/3219819.3219890}
\showDOI{\tempurl}


\bibitem[\protect\citeauthoryear{Zeng, Zhou, Srivastava, Kannan, and
  Prasanna}{Zeng et~al\mbox{.}}{2020}]%
        {graphsaint-iclr20}
\bibfield{author}{\bibinfo{person}{Hanqing Zeng}, \bibinfo{person}{Hongkuan
  Zhou}, \bibinfo{person}{Ajitesh Srivastava}, \bibinfo{person}{Rajgopal
  Kannan}, {and} \bibinfo{person}{Viktor Prasanna}.}
  \bibinfo{year}{2020}\natexlab{}.
\newblock \showarticletitle{{GraphSAINT}: Graph Sampling Based Inductive
  Learning Method}. In \bibinfo{booktitle}{\emph{International Conference on
  Learning Representations}}.
\newblock
\urldef\tempurl%
\url{https://openreview.net/forum?id=BJe8pkHFwS}
\showURL{%
\tempurl}


\bibitem[\protect\citeauthoryear{Zheng, Ma, Wang, Zhou, Su, Song, Gan, Zhang,
  and Karypis}{Zheng et~al\mbox{.}}{2020}]%
        {zheng2020distdgl}
\bibfield{author}{\bibinfo{person}{Da Zheng}, \bibinfo{person}{Chao Ma},
  \bibinfo{person}{Minjie Wang}, \bibinfo{person}{Jinjing Zhou},
  \bibinfo{person}{Qidong Su}, \bibinfo{person}{Xiang Song},
  \bibinfo{person}{Quan Gan}, \bibinfo{person}{Zheng Zhang}, {and}
  \bibinfo{person}{George Karypis}.} \bibinfo{year}{2020}\natexlab{}.
\newblock \bibinfo{title}{DistDGL: Distributed Graph Neural Network Training
  for Billion-Scale Graphs}.
\newblock
\newblock
\showeprint[arxiv]{2010.05337}~[cs.LG]


\end{thebibliography}

\end{document}